\documentclass[10pt,journal,compsoc]{IEEEtran}
\usepackage{times}
\usepackage{epsfig}
\usepackage{graphicx}
\usepackage{amsmath}
\usepackage{amssymb}
\usepackage{bm}
\usepackage{url}
\usepackage{cite}
\usepackage{enumerate}
\usepackage{hyperref}
\usepackage{amsfonts}
\usepackage{amstext}
\usepackage{multirow}
\usepackage{microtype}
\usepackage{booktabs}
\usepackage{subfigure}
\usepackage{rotating}
\usepackage{pifont}
\usepackage{indentfirst}
\usepackage[american]{babel}
\usepackage{makecell}

\newcommand{\etc}{\textit{etc}}
\newcommand{\etal}{\textit{et al}.}
\newcommand{\ie}{\textit{i}.\textit{e}.}
\newcommand{\eg}{\textit{e}.\textit{g}.}

\newcommand{\cmark}{\checkmark} 
\newcommand{\xmark}{\ding{55}} 

\newtheorem{definition}{Definition}

\begin{document}
%
\title{Data-Free Knowledge Transfer: A Survey}
%
%
%
%

\author{Yuang~Liu,
        Wei~Zhang,~\IEEEmembership{Member,~IEEE,}
        Jun~Wang, 
        and~Jianyong~Wang,~\IEEEmembership{Fellow,~IEEE}
\IEEEcompsocitemizethanks{\IEEEcompsocthanksitem Yuang Liu, Wei Zhang and Jun Wang are with the School of Computer Science and Technology, East China Normal University, Shanghai 200062, China (e-mail: frankliu624@gmail.com, zhangwei.thu2011@gmail.com, wongjun@gmail.com).
\IEEEcompsocthanksitem Jianyong Wang is with the Department of Computer Science and Technology, Tsinghua University, Beijing 100086, China, and also with the Jiangsu Collaborative Innovation Center for Language Ability, Jiangsu Normal University, Xuzhou 221009, China (e-mail: jianyong@tsinghua.edu.cn).
}
}

%
%

\markboth{Journal of \LaTeX\ Class Files,~Vol.~00, No.0, December~2021}%
{Liu \MakeLowercase{\textit{et al.}}: Data-Free Knowledge Transfer: A Survey}
%



\IEEEtitleabstractindextext{
\begin{abstract}
In the last decade, many deep learning models have been well trained and made a great success in various fields of machine intelligence, especially for computer vision and natural language processing. 
To better leverage the potential of these well-trained models in intra-domain or cross-domain transfer learning situations, knowledge distillation (KD) and domain adaptation (DA) are proposed and become research highlights.
They both aim to transfer useful information from a well-trained model with original training data.
However, the original data is not always available in many cases due to privacy, copyright or confidentiality. 
Recently, the data-free knowledge transfer paradigm has attracted appealing attention as it deals with distilling valuable knowledge from well-trained models without requiring to access to the training data. 
In particular, it mainly consists of the data-free knowledge distillation (DFKD) and source data-free domain adaptation (SFDA).
On the one hand, DFKD aims to transfer the intra-domain knowledge of original data from a cumbersome teacher network to a compact student network for model compression and efficient inference. 
On the other hand, the goal of SFDA is to reuse the cross-domain knowledge stored in a well-trained source model and adapt it to a target domain. 
In this paper, we provide a comprehensive survey on data-free knowledge transfer from the perspectives of knowledge distillation and unsupervised domain adaptation, to help readers have a better understanding of the current research status and ideas. 
Applications and challenges of the two areas are briefly reviewed, respectively. 
Furthermore, we provide some insights to the subject of future research. 
\end{abstract}

}

\maketitle

\IEEEdisplaynontitleabstractindextext

%
\IEEEpeerreviewmaketitle

\IEEEraisesectionheading{\section{Introduction}\label{sec:intro}}

\IEEEPARstart{W}{ith} the renaissance of deep learning, deep neural networks (DNN) have made significant progress in various fields of artificial intelligence, including computer vision (CV)~\cite{krizhevsky2012imagenet} and natural language processing (NLP)~\cite{goldberg2016primer}. In particular, the computer vision community has developed numerous applications of deep convolutional neural networks (such as image classification~\cite{xie2017aggregated}, object detection~\cite{ren2017faster}, semantic segmentation~\cite{chen2017rethinking}, \etc.), and greatly promoted the prosperity of deep learning. From LeNet~\cite{lecun1998gradient}, AlexNet~\cite{krizhevsky2012imagenet} to ResNet~\cite{he2016deep} and DenseNet~\cite{huang2017densely}, the remarkable success of the deep neural networks mainly depends on the over-parameterized architectures and the large-scale annotated training data. In practice, the applications of DNNs may bear up two problems: 1) the cumbersome models are impossible to be deployed on mobile devices with constrained storage and computation, such as autonomous driving cars~\cite{treml2016speeding} and real-time facial recognition systems~\cite{suk2014real}; 2) the entire labeled dataset is unavailable for training due to the unaffordable cost of annotations, for example, pixel-level annotations for semantic segmentation. 

\begin{figure}[!t]
\centering
\subfigure[KD]{
  \includegraphics[width=0.47\linewidth]{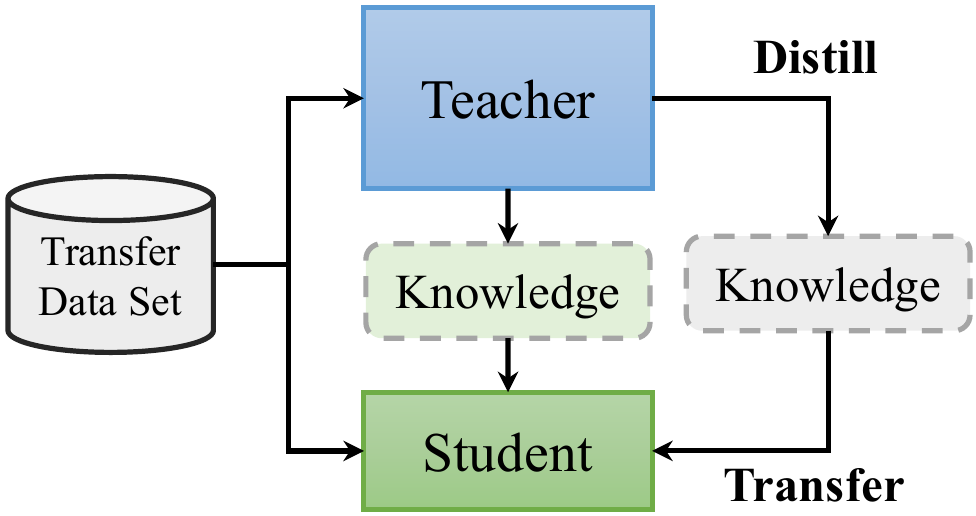}
}
\subfigure[UDA]{
  \includegraphics[width=0.47\linewidth]{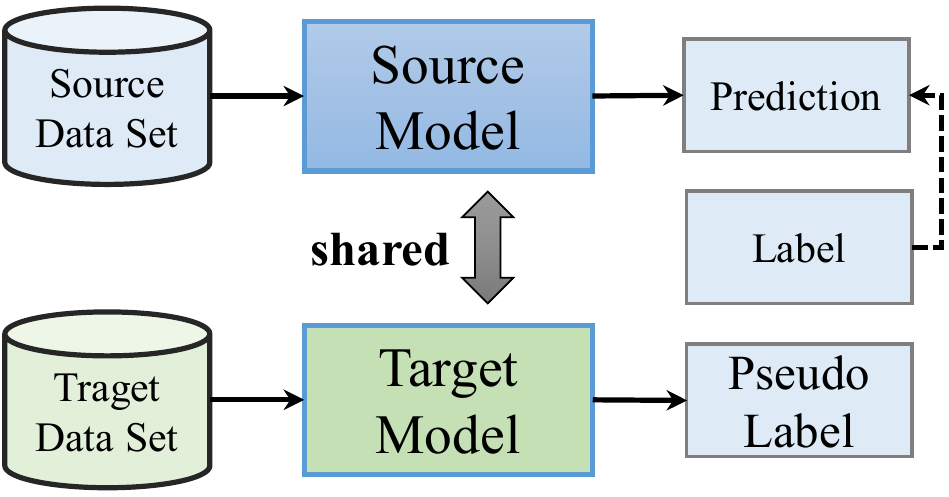}
}
\caption{Overview of knowledge distillation (KD) and unsupervised domain adaptation (UDA). }
\label{fig:intro1}
\end{figure}

\begin{figure}[b]
\centering
\subfigure[DFKD]{
  \includegraphics[width=0.47\linewidth]{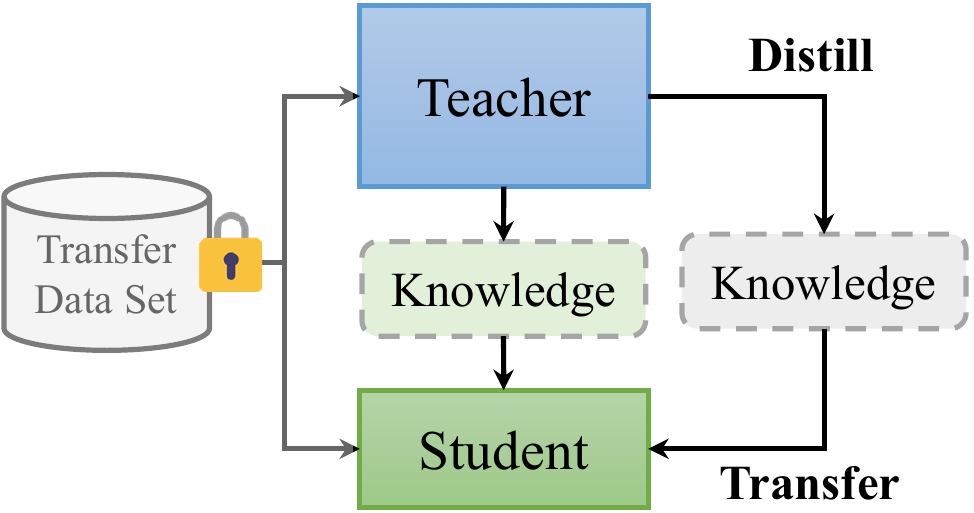}
}
\subfigure[SFDA]{
  \includegraphics[width=0.47\linewidth]{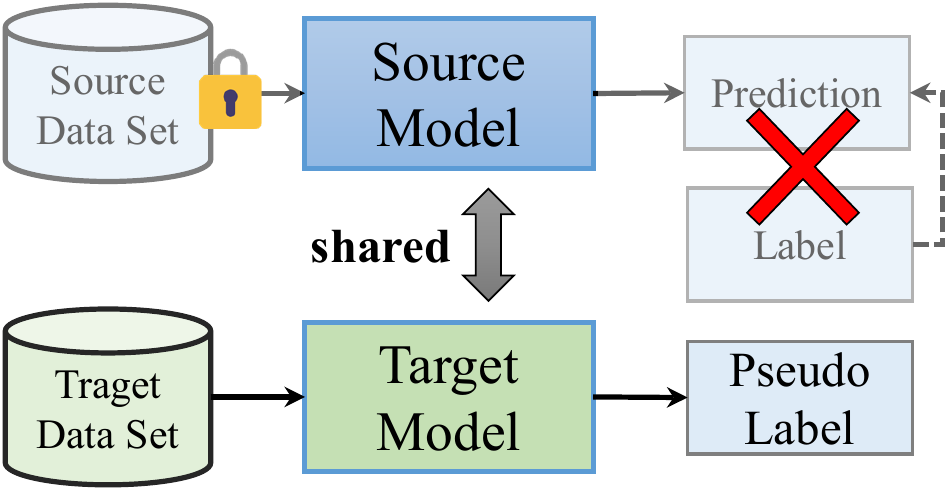}
}
\caption{Overview of data-free knowledge distillation (DFKD) and source-free domain adaptation (SFDA).}
\label{fig:intro2}
\end{figure}

\begin{figure*}[!t]
  \centering
  \includegraphics[width=.9\linewidth]{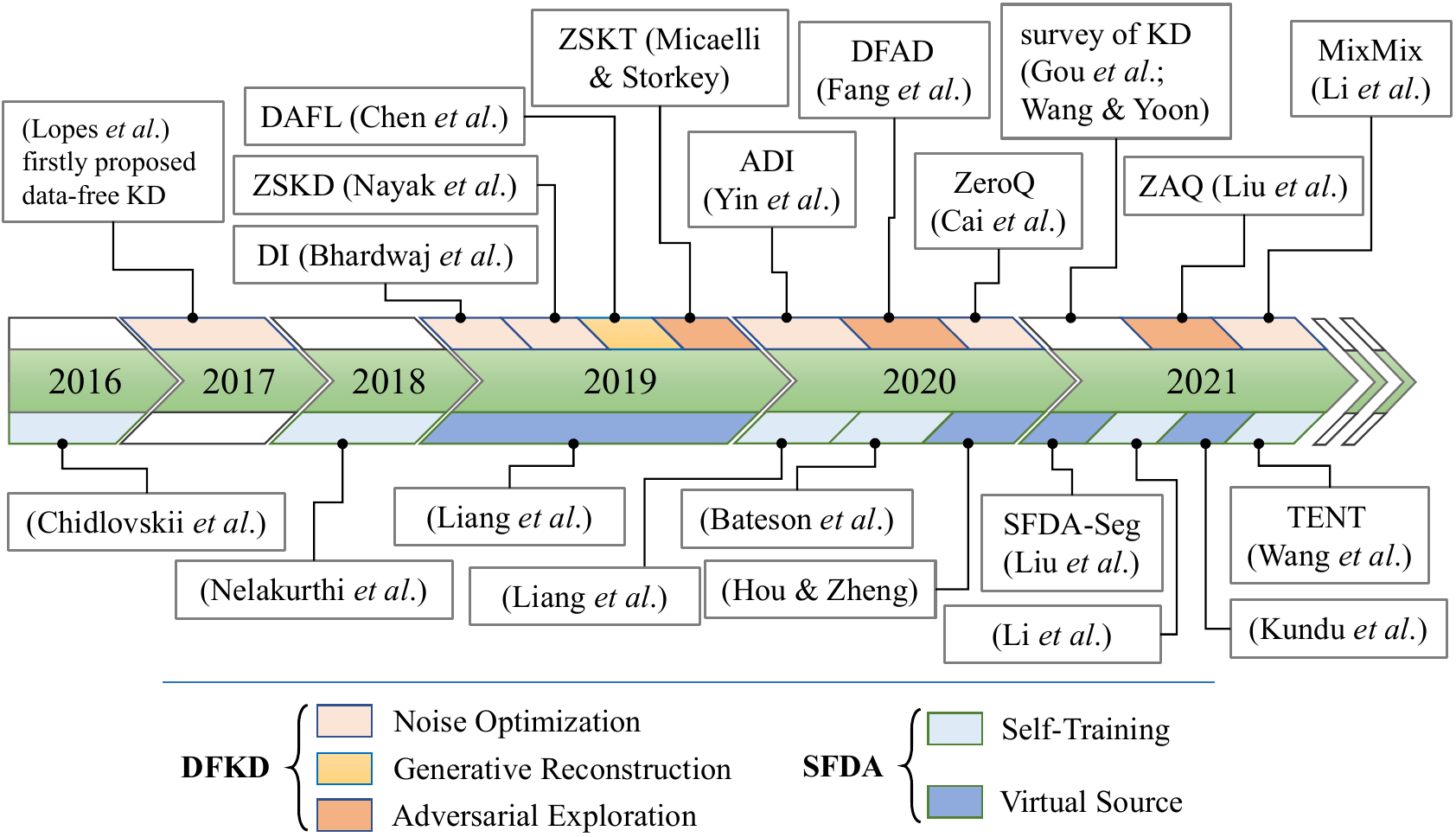}
  \caption{Timeline of data-free knowledge transfer from 2016 to 2021. The upper stream refers to the development of DFKD, while the lower stream represents the history of SFDA. The corresponding sub-categories are marked in different colors.}
  \label{fig:history}
\end{figure*}

To cope with the problem of deep model deployment, some works focus on reducing the storage and computation costs by model compression~\cite{bucilua2006model}, including pruning~\cite{li2016pruning}, quantization~\cite{zhou2018adaptive} and knowledge distillation~\cite{hinton2015distilling}. Knowledge distillation (KD)~\cite{hinton2015distilling} is a popular model compression way by transferring valuable information from a cumbersome teacher network to a compact student network. As a general teacher-student knowledge transfer framework shown in Fig.~\ref{fig:intro1}(a), it can be combined with other model compression methods without any specific design~\cite{polino2018model,kim2019qkd}. The student network is supposed to mimic the well-trained teacher network with the training data as input, which is very similar to the learning scheme of human. Most of the distillation methods are implemented to distill and transfer knowledge from the intermediate feature maps or predictions of the teacher network. 
In regard to model compression, the recent rapid development of knowledge distillation has a tremendous impact on semi-supervised learning~\cite{tang2021humble,chen2021semi}, incremental learning~\cite{chen2019new,dong2021few}, privacy protection~\cite{luo2018graph,cha2020proxy}, \etc. 

In addition to cumbersome network architectures, the high-cost annotations of large-scale datasets also limit the applications of deep learning. For instance, it takes about 90 minutes to manually annotate a Cityscapes~\cite{cordts2016cityscapes} image for semantic segmentation. An intuitive approach to address this issue is to leverage the specific knowledge from a related domain (source domain) for the considered target domain, which is inspired by human beings' study capabilities. 
Domain adaptation (DA)~\cite{zhuang2020comprehensive} is a promising transfer learning paradigm as shown in Fig.~\ref{fig:intro1}(b).
It aims to transfer knowledge from a source domain to a target domain, avoiding labor-intensive data annotations. 
According to the annotation rate of target domain data, domain adaptation can be further categorized into unsupervised DA (UDA), semi-supervised DA, and weak-supervised DA. Actually, only UDA methods entirely avoid the annotation cost, and we mainly consider the UDA setting in this paper. 

In summary, knowledge distillation and domain adaptation are two major research topics for transferring valuable knowledge from a well-trained deep neural network to an intra-domain or cross-domain network. 
The above-mentioned methods are all data-driven and rely on the premise that the original data or source data is accessible for distillation or domain adaptation. However, the original training data is unavailable in many practical cases due to privacy or copyright reasons. For example, there exist numerous pre-trained deep learning models~\cite{devlin2018bert,he2016deep,ren2017faster,chen2017rethinking} released by some well-known communities~\cite{paszke2019pytorch,wu2019detectron2,mmseg2020,mmdetection}.
But not all training data of them is accessible for compressing the pre-trained models or adapting them to the new domain.
Moreover, the medical or facial data is inaccessible for public or third-party institutions, because it involves the privacy of patients or users.
As such, how to transfer knowledge only with a well-trained model (no training data is available) becomes a new research topic.
It is summarized as ``Data-Free Knowledge Transfer (DFKT)'' shown in Fig.~\ref{fig:intro2}.
In particular, DFKT also involves two main research areas: (1) the knowledge distillation methods without training data are called Data-Free Knowledge Distillation (DFKD); (2) the domain adaptation methods without source data are called source data-free (or Source-Free) Domain Adaptation (SFDA). 
With the only available well-trained model (named as teacher in DFKD and source model in SFDA), DFKD aims to distill and transfer the original information of training dataset to a compact student model, while SFDA aims to query and explore the cross-domain knowledge through target data. In other words, DFKD transfers intra-domain knowledge between two models, while SFDA transfers cross-domain knowledge with a architecture-sharing model. 

\begin{figure*}[!t]
  \centering
  \includegraphics[width=.85\linewidth]{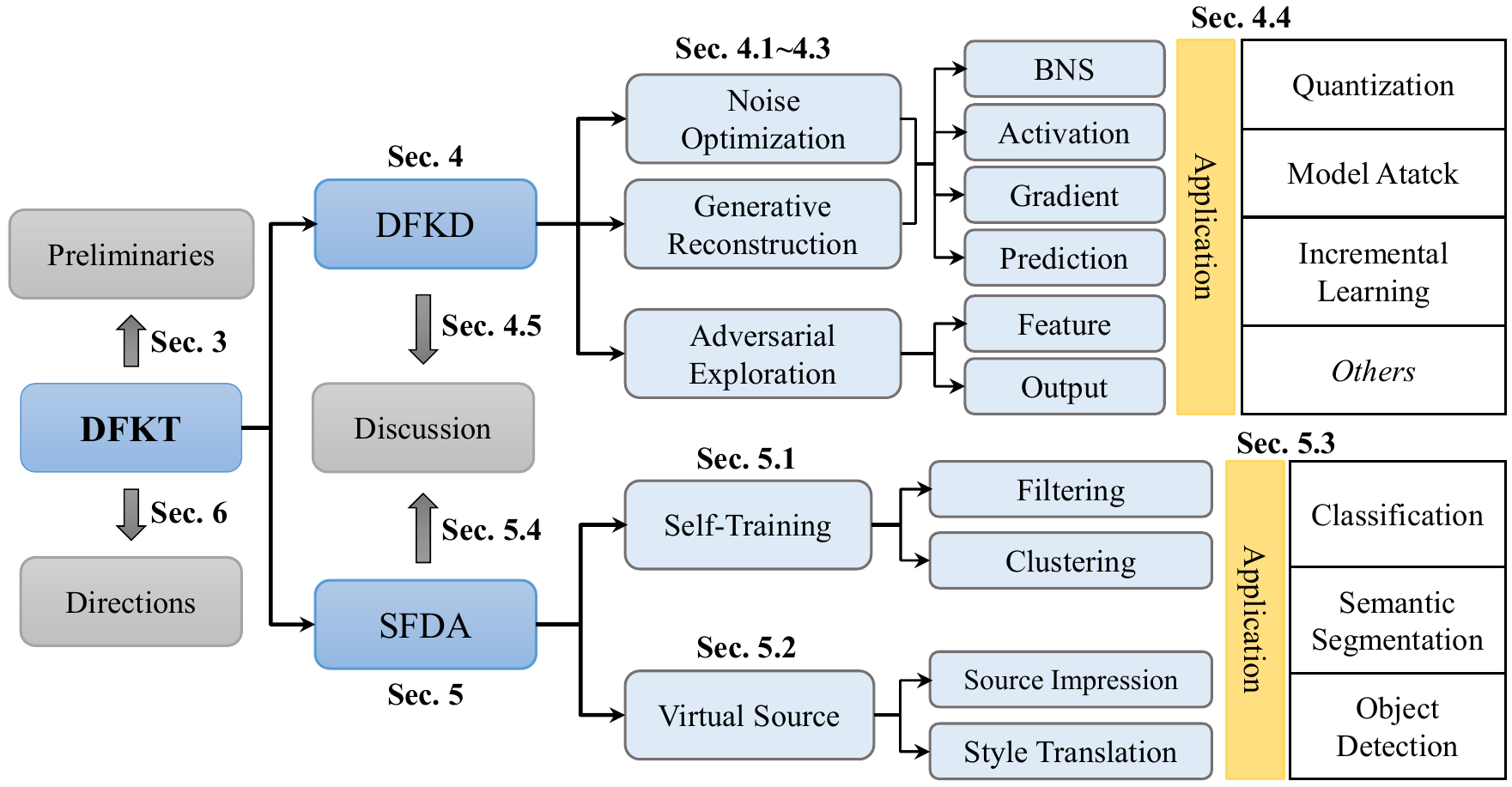}
  \caption{The structural taxonomy for data-free knowledge transfer. The survey is organized following the hierarchical structure. 
  }
  \label{fig:taxonomy}
\end{figure*}

In recent several years, the data-free knowledge transfer paradigm has attracted appealing attention in deep learning for various fields, especially computer vision (including image classification~\cite{nelakurthi2018source,nayak2019zero,yin2020dreaming}, object detection~\cite{saltori2020sf,chawla2021data,li2021free} and super-resolution~\cite{zhang2021data}). Fig.~\ref{fig:history} illustrates the timeline of data-free knowledge transfer. We depict the development of DFKD and SFDA on the upper and lower stream, respectively. Lopes~\etal~\cite{lopes2017data} firstly propose data-free knowledge distillation for DNNs in 2016.
It leverages summaries of the activations of a network to reconstruct its training set. With the boom of generative adversarial networks, some generative DFKD methods~\cite{nayak2019zero,chen2019data,micaelli2019zero,fang2020data} have been springing up like mushrooms since 2019, which attempt to synthesize alternative samples for knowledge transfer. 
And, some works follow \cite{lopes2017data} and utilize summaries of activations~\cite{bhardwaj2019dream} or batch normalization statistic (BNS)~\cite{yin2020dreaming,li2021learning} to recover the original image data from noises. Moreover, two comprehensive surveys of knowledge distillation~\cite{wang2021knowledge,gou2021knowledge} are published in 2021. 
As for SFDA, Chidlovskii~\etal~\cite{chidlovskii2016domain} present the pioneering work towards this direction. From 2018 to 2020, researchers mainly pay attention to source-free domain adaptation on classification~\cite{nelakurthi2018source,liang2020we,hou2020source}. The SFDA algorithms for semantic segmentation~\cite{liu2021source,wang2021tent} and object detection~\cite{saltori2020sf,li2021free} have been developed since 2020. There is no doubt that more and more researches on DFKT will be published in the future. 

Although traditional data-driven knowledge transfer has been a longstanding challenge in computer vision and achieved great success on model compression and cost reduction of data annotation, most of the works overlook the problems of data privacy and commercial copyright that grab increasing attention. 
Some researchers have presented comprehensive and detailed reviews on traditional data-driven knowledge distillation~\cite{wang2021knowledge,gou2021knowledge,alkhulaifi2021knowledge} and domain adaptation~\cite{wang2018deep,kouw2019review,zhuang2020comprehensive,csurka2021unsupervised}, in which the DFKD or SFDA is just simply illustrated as the tip of the iceberg. 
However, as DFKT matures, more and more relevant studies have been conducted, making it difficult for both the research and industrial communities to keep up with the pace of new progress. 
In view of this, a survey of the existing works is an urgent need and can be conducive to the communities. 
In this survey, we focus on categorizing and analyzing existing DFKD and SFDA methods under a unified data-free knowledge transfer framework. We discuss data-free knowledge distillation and source-free domain adaptation separately, while connect and compare them in terms of data reconstruction algorithms and knowledge transfer strategies. 
To give a straightforward understanding, we hierarchically categorize the DFKD and SFDA by their implementations, which is shown in Fig.~\ref{fig:taxonomy} and demonstrates the organization of our survey.
To summarize, our contributions are three-folds: 
\begin{itemize}
\item We give a systematic overview of data-free knowledge transfer, including taxonomy, definitions, two families of methods for DFKD and SFDA, and various applications. To our best knowledge, this is the first survey for DFKT. 
\item We provide a novel taxonomy that connects and unifies the data-free knowledge distillation and source-free domain adaptation from the perspective of intra- and cross-domain knowledge transfer. 
\item The advantages or challenges of each family of methods are roundly summarized, and some promising research directions are analyzed. 
\end{itemize}

In what follows, Sec.~\ref{sec:rw} briefly describes the related research ares, including knowledge distillation and domain adaptation. Sec.~\ref{sec:overview} gives some fundamental notations and definitions. Sec.~\ref{sec:dfkd} reviews existing data-free knowledge distillation methods according to the data reconstruction strategies, and Sec.~\ref{sec:sfda} categories source-free domain adaptation methods into self-training and virtual source. Both application and discussion are illustrated in the two sections. The future directions in DFKT are discussed in Sec.~\ref{sec:frd}. Finally, Sec.~\ref{sec:con} concludes the paper.

\section{Related Research Area}
\label{sec:rw}

In this section, we briefly review the works on data-driven knowledge transfer, including knowledge distillation and domain adaptation two areas.

\subsection{Knowledge Distillation}
\label{sec:kd}

Knowledge distillation, proposed by Hinton~\etal~\cite{hinton2015distilling}, is a popular teacher-student learning architecture mainly for model compression.
It has been applied to computer vision (not limited to semantic segmentation~\cite{liu2019structured,wang2020intra}, pose estimation~\cite{zhang2019fast,weinzaepfel2020dope}, and object detection~\cite{deng2019relation,chen2021deep}), natural language processing~\cite{mukherjee2020tinymbert,liu2020fastbert}, and even recommender systems~\cite{TangW18}.
According to the forms of knowledge, the majority of current knowledge distillation approaches can be roughly divided into three categories, \ie, prediction distillation~\cite{hinton2015distilling,yang2019snapshot,zhang2019fast,zhang2020prime}, feature distillation~\cite{romero2015fitnets,zagoruyko2016paying,heo2019comprehensive,wang2019distilling,chung2020feature}, and relation distillation~\cite{park2019relational,liu2019knowledge,chen2020improving,liu2021exploring}. The prediction distillation seeks to transfer the implicit information of label distributions in the logits, while the feature distillation aims to utilize the intermediate feature maps to guide the student network. In addition, the relation distillation generally conducts structural relations among samples or distills contextual relations to help the student learning.  
According to the learning process, it can be categorized into self-distillation~\cite{hou2019learning,kim2021self}, offline/online distillation~\cite{zhang2018deep,chen2020online}, ensemble distillation~\cite{park2019feed,liu2020adaptive,du2020agree} and so on. Hou~\etal~\cite{hou2019learning} propose an efficient lane detection algorithm where the network is supervised by the features distilled from itself, without an assistant teacher. Most of the KD methods are learned offline, meaning that the parameters of the teacher are fixed. On the contrary, online distillation learning~\cite{zhang2018deep,chen2020online} employs a trainable teacher and updates it synchronously with the student.
To distill rich and robust knowledge, some studies~\cite{park2019feed,liu2020adaptive,du2020agree} introduce multiple teachers to guide the student. 
Moreover, KD has been extended to some situations like semi-supervised learning~\cite{liu2021unbiased,cai2021exponential}, incremental learning~\cite{hu2021distilling,cheraghian2021semantic}, reinforcement learning~\cite{lai2020dual,fang2021universal}, \etc. 
It's worth nothing that all the above KD methods are data-driven ones. Since the flourish of the KD methods, two surveys~\cite{wang2021knowledge,gou2021knowledge} have been published recently, where the data-free KD is briefly illustrated as a subsection. 
By contrast, we focus on reviewing data-free KD specially and connecting it with source-free DA in this survey.

\subsection{Domain Adaptation}

Domain adaptation is a longstanding transfer learning paradigm with source-trained models for avoiding expensive annotation of target data. Most DA methods tackle the domain-shift problem between the labeled source domain and unlabeled target domain by aligning the feature distributions, which can be divided into three categories. 
The first category of methods~\cite{zou2018unsupervised,chen2019maxsquare,pan2020unsupervised} seek to filter and select confident pseudo-labels of target data for self-supervised training. Chen~\etal~\cite{chen2019maxsquare} propose a new supervised MaxSquare loss for the target adaptation. IntraDA~\cite{pan2020unsupervised} is a two-step self-supervised domain adaptation approach to minimize the inter-domain and intra-domain gap together. 
The second category of approaches~\cite{ganin2016domain,tsai2019domain,vu2019advent} focus on achieving domain alignment by adversarial training. DANN~\cite{ganin2016domain} is inspired by the theory on domain adaptation.
It suggests that, for effective domain transfer to be achieved, predictions must be made based on features that cannot discriminate between the training (source) and test (target) domains. Tsai~\etal~\cite{tsai2019domain} propose to learn discriminative feature representations of patches in the source domain by discovering multiple modes of patch-wise output distribution through the construction of a clustered space. 
The third category~\cite{hoffman2018cycada,li2019bidirectional,luo2020adversarial} tends to introduce an image translation network to transfer style between the source and target images, which can fit the domain shift. Hoffman~\etal~\cite{hoffman2018cycada} develop a model which adapts between domains using both generative image space alignment and latent representation space alignment. Luo~\etal~\cite{luo2020adversarial} propose to address the one-shot target DA issue by combining the style transfer module and task-specific module into an adversarial manner. 
The above traditional DA methods are all based on source data.
Several surveys~\cite{wang2018deep,kouw2019review,zhuang2020comprehensive,csurka2021unsupervised} have illustrated them in details. Differently, we focus on introducing and analyzing the source-free DA methods.

\section{Preliminaries}
\label{sec:overview}

We provide an overview and formulation of the data-free knowledge transfer framework in this section. To assist the formal description of each kind of approaches in the following section, some significant notations and definitions are given here. 

\subsection{Notation}

For convenience, a list of symbols and their definitions are shown in Tab.~\ref{tab:notation}. Note that $\mathcal{P}$ and $\mathcal{Q}$ share the same architecture but with different input domains in the DA case. 

\begin{table}[thbp]
  \centering
  \caption{Notations used in this paper.}
  \label{tab:notation}
  \begin{tabular}{l|l}
  \toprule
  \textbf{Symbol} & \textbf{Description} \\
  \midrule
  $\mathcal{P}$ & Teacher/Source model \\
  $\mathcal{Q}$ & Student/Target model \\
  $\mathcal{G}$ & Generator \\ 
  $\boldsymbol{z}$ & Noise vector \\
  $\tilde{X}$ & Alternative data set \\
  $\tilde{Y}$ & Label set corresponding to $\tilde{X}$ \\
  $X$ & Data set \\
  $Y$ & Label set corresponding to $X$ \\
  $H, W$ & Height and Width\\
  $\tilde{x}, \tilde{y}$ & Alternative sample and label \\
  $\delta$ & Softmax operation \\
  $\mu $ & Mean of BN layer \\
  $\sigma$ & Variance of BN layer \\
  $\mathcal{D}$ & Model discrepancy \\
  $\mathcal{L}$ & Loss function \\
  $\mathcal{R}$ & Regularization function \\
  $\Phi$ & Domain space \\
  $C$ & Number of classes \\
  $h$ & Softmax vector \\
  $\boldsymbol{f}$ & Feature vector \\
  $l$ & Distance function \\
  $p$ & Distribution function of probability \\
  $\mathbb{E}$ & Mathematical expectation \\
  $\mathcal{N}(m, v)$ & Gaussian distribution with mean $m$ and variance $v$ \\
  \bottomrule
  \end{tabular}
\end{table}

\subsection{Definition}

In this subsection, we give the definitions of data impression and alternative data, which are generally utilized in DFKT. The basic formulations of DFKT, DFKD, and SFDA are defined for an overview. 

\begin{definition}[DFKT]
  Given a well-trained model $\mathcal{P}$ on an inaccessible source domain $\Phi^s$ and a specified domain $\Phi^t$, data-free knowledge transfer utilizes the valuable knowledge implied in $\mathcal{P}$ to improve the performance of the new model $\mathcal{Q}$ on the specified domain $\Phi^t$.
\end{definition}
There are some key points to explain: 
(1) The source domain $\Phi^s$ and the specified domain $\Phi^t$ might be the same in intra-domain knowledge transfer.
And in domain transfer, the specified domain $\Phi^t$ corresponds to a target domain.
(2) The two models involved in the DFKT perform the same task, and the transfer learning only exists between models or between domains within the scope of our discussion. 
(3) In practice, a domain is often observed by a number of instances with/without the label information~\cite{zhuang2020comprehensive}. So a dataset (\eg, $X^t$) contains some representative instances of the corresponding domain (\eg, $\Phi^t$). 
The inaccessible domain means that the data of the source domain is unavailable. 

\begin{definition}[Data Impression]
  Given a model $\mathcal{P}$ pre-trained on an inaccessible source dataset $X^s$, the samples synthesized from $\mathcal{P}$ could contain similar visual content and have a same label space with the images in $X^s$ of the corresponding domain $\Phi^s$, which are defined to be the data impressions of $X^s$. 
\end{definition}
If a domain $\Phi^s$ can be observed by the instances in the dataset $X^s$, then a data impression set of $X^s$ can be regarded as another observation of $\Phi^s$. 

\begin{definition}[Alternative Data]
  Given a well-trained model $\mathcal{P}$ on an inaccessible source domain $\Phi^s$, the synthesized samples that can be employed to replace the original training data in $X^s$ and distill the domain knowledge of $\Phi^s$ in teacher-student architectures, are regarded as alternative data for DFKT. 
\end{definition}
From the definitions of data impressions and alternative data, we can see that the data impressions are a kind of alternative data. But the alternative data does not need to be similar to the original training data in terms of visual content. 

\begin{figure*}[!t]
\centering
\subfigure[Noise Optimization]{
  \includegraphics[width=0.31\linewidth]{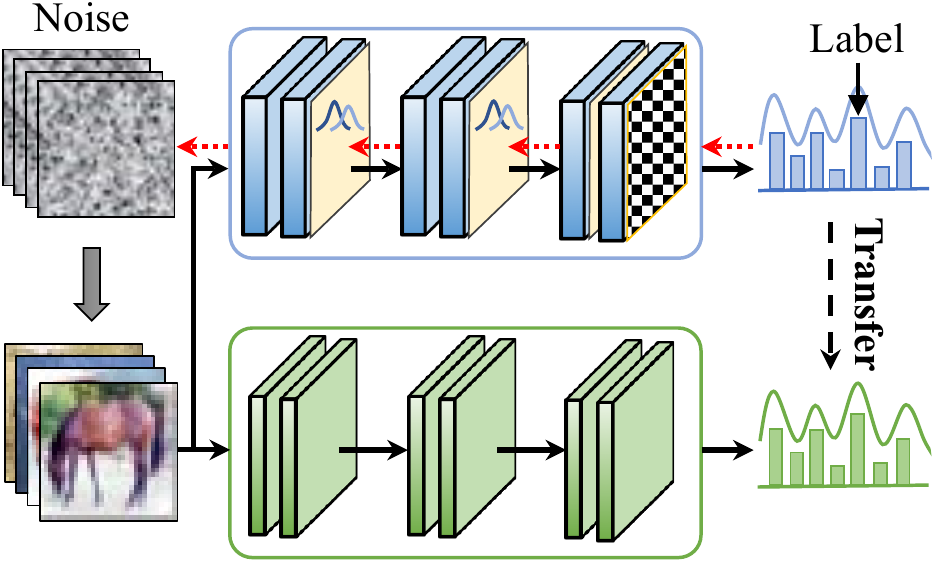}
}\quad
\subfigure[Generative Reconstruction]{
  \includegraphics[width=0.31\linewidth]{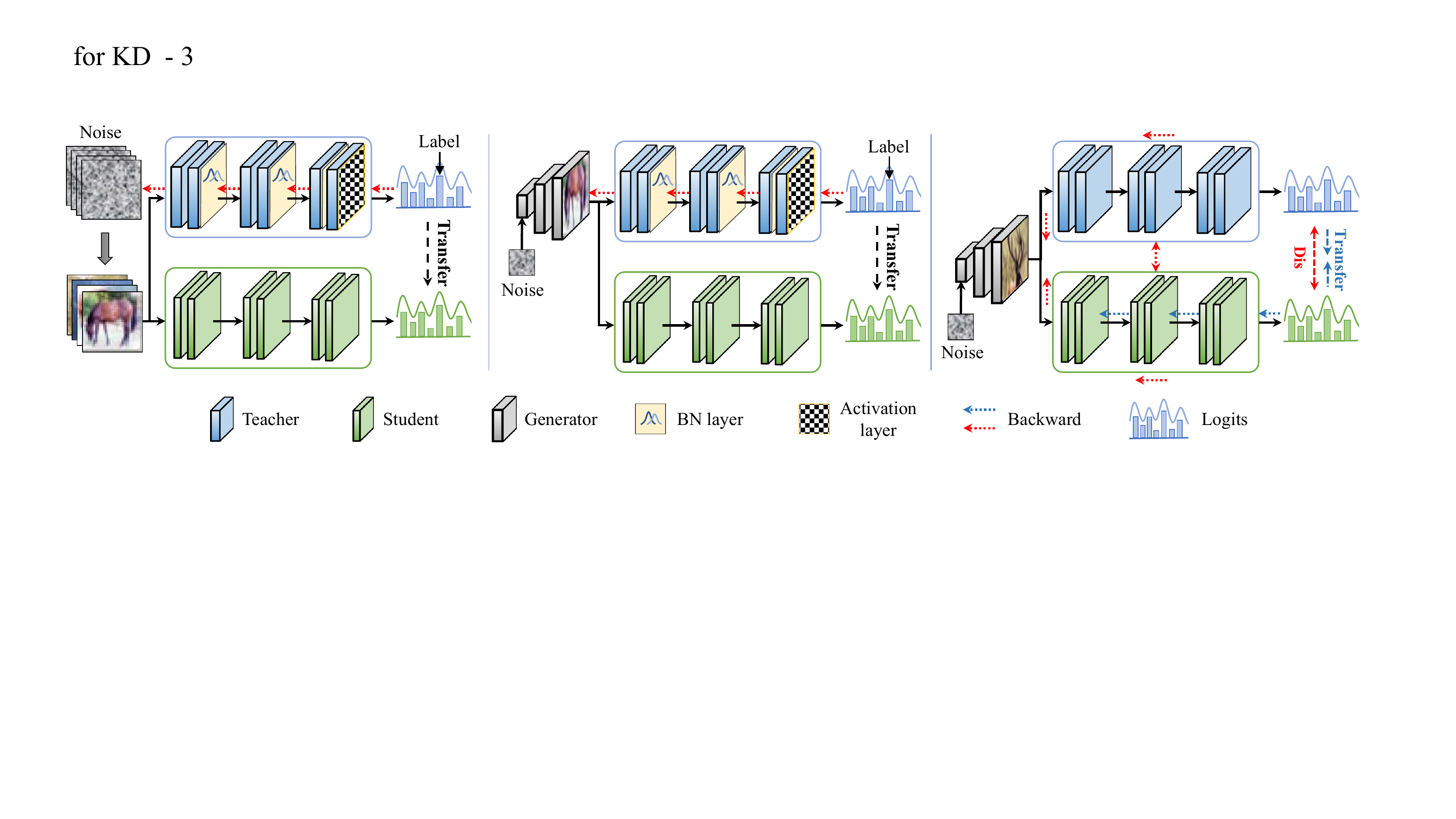}
}\quad
\subfigure[Adversarial Exploration]{
  \includegraphics[width=0.31\linewidth]{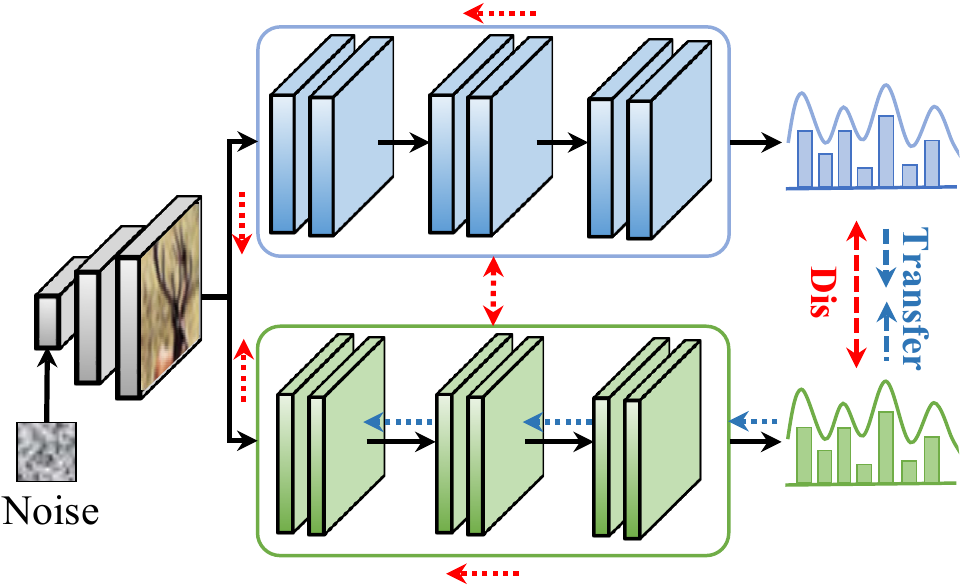}
}

\subfigure{
  \includegraphics[width=0.86\linewidth]{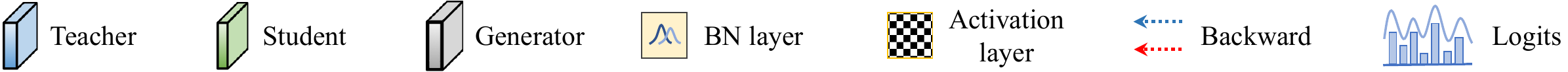}
}
\caption{Three kinds of data-free knowledge distillation frameworks. The legends of some components are listed for clarification. }
\label{fig:dfkd}
\end{figure*}

\begin{definition}[DFKD]
  As a subfield of DFKT, data-free knowledge distillation utilizes the knowledge of domain $\Phi^s$ implied in the well-trained model $\mathcal{P}$ to train another compressed model $\mathcal{Q}$ on $\Phi^s$, without accessing to the original training data $X^s$. 
\end{definition}

\begin{definition}[SFDA]
  As another subfield of DFKT, source-free domain adaptation transfers the valuable knowledge from the well-trained model $\mathcal{P}$ by only using the unlabeled target dataset $X^t$, hoping to improve the performance of $\mathcal{Q}$ on the target domain $\Phi^t$. 
\end{definition}
The models $\mathcal{P}$ and $\mathcal{Q}$ generally share the same model architecture in DA, but with input data from different domains.

\section{Data-Free Knowledge Distillation}
\label{sec:dfkd}

Aiming at transferring knowledge between two models under the teacher-student framework, the core of data-free knowledge distillation is to synthesize alternatives of the original training data. According to the synthesis algorithms of the alternative data, the major data-free knowledge distillation methods can be divided into three categories: 1) noise optimization, 2) generative reconstruction, and 3) adversarial exploration.
The major thought of them are briefly depicted in Fig.~\ref{fig:dfkd}. 
In this section, we introduce these data-free knowledge distillation approaches and their applications in various fields, \ie, quantization \& pruning, incremental learning, model attack, to name a few. 
The advantages and challenges of different DFKD approaches are discussed in the end.

\subsection{Noise Optimization}
\label{sec:np}
In DFKD, the major requirement is the availability of a well-trained model $\mathcal{P}$, which stores the prior information of the original training data $X^s$ and the corresponding domain $\Phi^s$.
To achieve distillation or transfer learning, a straightforward solution is to reconstruct enough realistic training data or synthesize data impressions of the domain $\Phi^s$.
Noise optimization is such a type of methods.
In fact, it is a two-stage solution: the first step is data reconstruction and the second step is knowledge transfer, as shown in Tab.~\ref{fig:dfkd}(a).
In the first stage, it usually samples a noise from a Gaussian distribution as input and optimize it with gradient descent iteratively until meeting certain constraints.
With an alternative dataset $(\tilde{X}, \tilde{Y})$ consisting of enough reconstructed data, the knowledge transfer stage can follow the traditional data-driven KD approaches~\cite{wang2021knowledge,gou2021knowledge}.

Assuming that a data impression $\tilde{x}\in \mathbb{R}^{H\times W\times 3}$ is randomly initialized by the Gaussian distribution $\mathcal{N}(m, v)$, the objective function for optimizing the data impression is defined as 
\begin{equation}
  \tilde{x}^{*} = \arg\min_{\tilde{x}}\mathcal{R}(\tilde{x} | \mathcal{P}) \,,
\end{equation}
where $\mathcal{R}$ is a regularization function about $\mathcal{P}$, providing a prior constraint for $\tilde{x}$. 
As for the label $\tilde{y}$ of the data impression $\tilde{x}$, a simple approach is to directly use the label predicted by $\mathcal{P}(\tilde{x})$. 
If we want to control the class distribution of data impressions, the label $\tilde{y}$ can be pre-defined and constrain the noise optimization by an additional cross-entropy loss:
\begin{equation}
  \tilde{x}^{*} = \arg\min_{\tilde{x}} \mathcal{R}(\tilde{x}, \mathcal{P}) + \mathcal{L}_{CE}(\mathcal{P}(\tilde{x}), \tilde{y})\,.
\end{equation}
Then the labeled data impressions build up the alternative dataset $(\tilde{X}, \tilde{Y})$, which is utilized in the second-stage knowledge transfer learning, formulated as follows:
\begin{equation}
  \label{eq:kd}
  \min_{\mathcal{Q}} \sum_{(\tilde{x}, \tilde{y})}^{(\tilde{X}, \tilde{Y})} \mathcal{L}_{CE}(\mathcal{Q}(\tilde{x}), \tilde{y}) + \mathcal{L}_{KL}(\delta(\mathcal{Q}(\tilde{x})), \delta(\mathcal{P}(\tilde{x}))) \,,
\end{equation}
where $\mathcal{L}_{KL}$ computes the Kullback-Leibler (KL) divergence between its two inputs and $\delta(\cdot)$ is the softmax operation. 

With these insights, we can see that the core of noise optimization is to devise a suitable regularization function $\mathcal{R}$ for distilling prior knowledge.
To achieve this, most existing studies leverage four kinds of prior information from different components of deep neural networks, \ie, activation, prediction, batch normalization statistics (BNS), and gradient. 

\textbf{1) Activation}. As a pioneering study of DFKD, Lopes~\etal~\cite{lopes2017data} argue that the summary of the activations of a network on its training set can reflect the prior distribution of the training data. Thus, the activation records (\ie, means and covariance matrices) of each layer can be stored alongside the model after training and used as metadata for reconstructing data impressions. Given a neural network representation $\phi$ and an initial network activation $\phi_{0}$, the data impression $\tilde{x}$ can be iteratively optimized by 
\begin{equation}
  \tilde{x}^{*} = \arg\min_{\tilde{x}} l(\phi(\tilde{x}), \phi_0) \,,
\end{equation}
where $l(\cdot, \cdot)$ is a specified distance function. To reduce the privacy concerns, 10\% real CIFAR-10 images, instead of the whole dataset, are passed through a pre-trained teacher model to obtain metadata~\cite{bhardwaj2019dream}. 
Specifically, the cluster centroids of activation vectors are computed and stored as metadata. 
Although these methods do consider the case of knowledge distillation in the absence of training data, the metadata still depends on the training data.
Therefore, they are not purely data-free approaches. Moreover, the experiments demonstrate that the activation records cannot represent complex information about the training data, which hinders the methods to be applied to large models and high-resolution images. 

\textbf{2) Prediction.} ZSKD~\cite{nayak2019zero} is actually the first complete DFKD method.
It does not access to the original training data or depend on any metadata like \cite{lopes2017data,bhardwaj2019dream}. 
Instead, it presents a sample extraction mechanism via modelling the softmax space (prediction) as a Dirichlet distribution and crafting data impressions from the parameters of the teacher model $\mathcal{P}$. 
It is worth noting that the softmax vector $\tilde{h}$ of the data impression $\tilde{x}$ is modeled and sampled before the noise optimization.
Hence the label $\tilde{y}$ is pre-defined by one-hot operation in this method. With the pre-defined softmax vector $\tilde{h}$, the data impression $\tilde{x}$ is simply optimized via cross-entropy loss 
\begin{equation}
  \tilde{x}^{*} = \arg\min_{\tilde{x}}\mathcal{L}_{CE}(\mathcal{P}(\tilde{x}), \tilde{h}) \,.
\end{equation}
This work investigates the effect of the size of alternative dataset and observes that the performance increases with size.
And the initial performance (with a smaller alternative dataset) reflects the complexity of the task. 

\textbf{3) BNS.} To derive artistic effects on natural images, DeepDream~\cite{mordvintsev2015inceptionism} proposes an image prior term to steer the generated images away from unrealistic images that are classified correctly but possess no discernible visual information:
\begin{equation}
  \mathcal{R}_{\text{img}}(\tilde{x}) = \mathcal{R}_{\text{TV}}(\tilde{x}) + \mathcal{R}_{l_2}(\tilde{x}) \,, 
\end{equation}
where $\mathcal{R}_{\text{TV}}$ and $\mathcal{R}_{l_2}$ penalize the total variance and $l_2$ norm of $\tilde{x}$.
To gain more realistic images, DeepInversion~\cite{yin2020dreaming} extends the DeepDream with a new feature distribution regularization term by utilizing BNS stored in the batch normalization layers. 
Specifically, it assumes the batch-wise statistics in each layer are Gaussian distributed.
Then the running mean $\mu_l$ and running variance $\sigma^2_l$ stored in the $l$-th BN layer can be utilized as a feature prior of the original data. 
It minimizes the distance between the statistics of feature maps for $\tilde{x}$ and the BNS prior, which can be formulated as 
\begin{equation}
  \mathcal{R}_{\text{BNS}}(\tilde{x}) = \sum_{l} \left( \| \tilde{\mu}_l(\tilde{x}) -\mu_l \|^2_2 + \| \tilde{\sigma}^2_l(\tilde{x}) - \sigma^2_l \|^2_2 \right) \,,
\end{equation}
where $\tilde{\mu}_l(\tilde{x})$ and $\tilde{\sigma}^2_l(\tilde{x})$ are the estimated mean and variance of the feature maps corresponding to the $l$-th convolutional layer. 
The total objective of DeepInversion can be formulated as
\begin{equation}
  \tilde{x}^{*} = \arg\min_{\tilde{x}} \mathcal{R}_{\text{BNS}}(\tilde{x}) + \mathcal{R}_{\text{img}}(\tilde{x}) \,.
\end{equation}
The image prior $\mathcal{R}_{\text{img}}$ is usually used as an adjunct to the BNS regularization, so they are combined in this paper. 
Inspired by the multi-teacher KD algorithms~\cite{du2020agree,liu2020adaptive}, recently MixMix~\cite{li2021learning} introduces feature mixing and data mixing techniques to synthesize data impressions via BNS of multiple teacher models. The feature mixing can absorb the knowledge from widespread pre-trained architectures and generalize the data impressions to many models and applications. This is very different from the previous DFKD methods with noise optimization. Due to the contribution to high-quality and realistic image prior, BNS has been the most popular way to synthesize data impressions in various DFKD algorithms and applications, which will be illustrated in Sec.~\ref{sec:app1}. 

\textbf{4) Gradient.} DLG~\cite{zhu2019deep} firstly demonstrates that it is possible to obtain the private training data from the publicly shared gradients in modern multi-node learning systems (\eg, distributed training and federated learning). \cite{jonas2020inverting} and \cite{yin2021see} both propose advanced gradient inversion algorithms to recover images in federated learning. 
To be concrete, they recover the privacy data by minimizing the distance between the gradients of the current node and the public node.
However, the techniques cannot be applied in common teacher-student DFKD architectures, where no public gradients of training data are available. 
To relieve this issue, some researchers have been starting to study KD in federated learning systems~\cite{cha2020proxy,zhu2021data}, to which the gradient inversion algorithms could make a significant contribution. 

Though realistic data impressions or alternative dataset can be collected via iterative noise optimization for knowledge distillation, this kind of DFKD methods has two weaknesses: 1) every data impression needs to go through hundreds or thousands of gradient descent, causing unbearable time consumption; 2) unless it is verified in the second stage, the quality of the alternative dataset collected in the first stage cannot be estimated. By contrast, generative reconstruction based DFKD methods could address the above issues, which is introduced later.

\subsection{Generative Reconstruction}

Generative Adversarial Networks (GAN)~\cite{goodfellow2014generative} have received wide attention in the image synthesis field for their potential to learn high-dimensional and complex real data distribution.
Typically, GAN consists of a generator $\mathcal{G}$ and a discriminator $\mathcal{D}$. $\mathcal{G}$ is expected to synthesize desired data and fool $\mathcal{D}$ while $\mathcal{D}$ is trained to distinguish if the data is produced by $\mathcal{G}$. 
As aforementioned, the distribution of the original data $x^s$ can be estimated with some prior information from the well-trained model $\mathcal{P}$, \eg, prediction, BNS, activation, and gradient. 
So a generator $\mathcal{G}$ can be introduced to estimate the distribution of $x^s$ and synthesize data impressions, which is named generative reconstruction based DFKD and depicted in Fig.~\ref{fig:dfkd}(b). To be specific, given a noise vector $\boldsymbol{z}$ as a low-dimensional representation, the generator $\mathcal{G}$ maps $\boldsymbol{z}$ to the desired data impression $\tilde{x}$.
It is supposed to satisfy the regularization constraint $\mathcal{R}(\tilde{x}|\mathcal{P})$ for prior knowledge. 
Four kinds of regularization $\mathcal{R}(\tilde{x}|\mathcal{P})$ have been defined in Sec.~\ref{sec:np}, and can also be utilized in generative reconstruction methods. 

With these insights, we can collect the data impressions by a generative approach,
\begin{equation}
  \label{eq:gen}
  \tilde{x} = \mathcal{G}(\boldsymbol{z}) \,, \quad \boldsymbol{z} \sim p_z(\boldsymbol{z}) \,,
\end{equation}
where $p_z(\boldsymbol{z})$ is the distribution of noise $\boldsymbol{z}$ (\eg, a Gaussian distribution $\mathcal{N}(m, v)$). 
In the training procedure, the generator is continuously updated according to the training error produced by $\mathcal{R}(\tilde{x} | \mathcal{P})$, which is formulated as 
\begin{equation}
  \min_{\mathcal{G}} \mathbb{E}_{\boldsymbol{z} \sim p_z(\boldsymbol{z})} \mathcal{R}(\mathcal{G}(\boldsymbol{z}) | \mathcal{P}) \,.
\end{equation}
The label $\tilde{y}$ of the data impression $\tilde{x}$ can be obtained by the two ways illustrated in the Sec.~\ref{sec:np}. But if the label $\tilde{y}$ is pre-defined in a classification task, $\mathcal{G}$ can be optically extended to a conditional generator $\mathcal{G}(\boldsymbol{z}, \tilde{y})$~\cite{yoo2019knowledge} that synthesizes data impressions according to the label condition. Then, $\mathcal{G}$ is upgraded by a regularization loss and cross-entropy loss as
\begin{equation}
  \min_{\mathcal{G}} \mathbb{E}_{\boldsymbol{z} \sim p_z(\boldsymbol{z})} \left[ \mathcal{R}(\mathcal{G}(\boldsymbol{z}, \tilde{y}) | \mathcal{P}) + \mathcal{L}_{CE}(\mathcal{G}(\boldsymbol{z}, \tilde{y}), \tilde{y}) \right] \,.
\end{equation}
After each iteration of data generation, the batch of data impressions can be utilized for knowledge transfer with Eqn.~\ref{eq:kd}. 
Different from the noise optimization methods, the generative reconstruction based DFKD methods alternately executes data synthesis and knowledge transfer in each round of iteration, instead of collecting the whole alternative dataset to transfer. 
Therefore, each one or each batch of data impressions only needs one time of generator update.
This avoids numerous iterations with noise optimization. 
The following are some typical generative reconstruction based DFKD methods. 

DAFL~\cite{chen2019data} is the first effort to introduce generative reconstruction for DFKD.
Despite the cross-entropy loss function, they propose two regularization functions about activation and prediction to help data synthesis. And these two regularization functions are widely employed in the follow-up works~\cite{ye2020data,liu2021zero,fang2021contrastive}. Features extracted by the convolutional filters are supposed to express high-level and rich information about the input data. Since filters in the teacher model have been trained to extract intrinsic patterns in training data, feature maps tend to receive higher activation values if input images are real rather than some random representations.
Motivated by this, an \textbf{activation} regularization function is defined as follows:
\begin{equation}
  \mathcal{R}_{\text{act}} = - \| \boldsymbol{f}_P \|_1 \,,
\end{equation}
where $\boldsymbol{f}_P$ is the feature vector extracted by the teacher model $\mathcal{P}$. Besides, the realistic data is supposed to cause a low entropy value of the teacher's \textbf{prediction}. 
Given a softmax vector $\tilde{h} = (\tilde{h}^{(1)}, \tilde{h}^{(2)}, \cdots, \tilde{h}^{(C)})$ of $\tilde{x}$, the information entropy regularization is formulated as 
\begin{equation}
  \mathcal{R}_{\text{ie}} = - \frac{1}{C}\sum_i^C \tilde{h}^{(i)} \log(\tilde{h}^{(i)}) \,. 
\end{equation}
Ye~\etal~\cite{ye2020data} propose a data-free knowledge amalgamate strategy to craft a well-behaved multi-task student network from multiple single/multi-task teachers. They construct a group-stack generative adversarial networks to reconstruct data impressions from multiple teachers, with the three regularization functions adopted in DAFL~\cite{chen2019data}. In fact, it can be considered as a multi-teacher/multi-task version of DAFL. Instead of employing multiple teachers, Luo~\etal~\cite{luo2020large} build an ensemble of generators to synthesize data impressions according to \textbf{BNS}. In DF-IKD~\cite{shah2021empirical}, the authors modify the knowledge distillation loss function and conduct an empirical study of data-free iterative knowledge distillation, \ie, namely DAFL. 

KegNet~\cite{yoo2019knowledge} adopts a conditional generator and an auxiliary decoder to extract the knowledge of a trained neural network for knowledge distillation without observable data. The generator creates artificial data and feeds it into the classifier and decoder. The fixed classifier produces the label distribution of each data point, and the decoder finds its hidden representation as a low-dimensional vector, \ie, the input noise. This work actually studies model attack or extraction problem, where only the data reconstruction is considered, without knowledge distillation procedure. 

Both \cite{haroush2020knowledge} and \cite{besnier2020dataset} introduce generative reconstruction method to synthesize data impressions via the \textbf{BNS} prior. In addition, \cite{haroush2020knowledge} utilizes an inception scheme, \ie, \textbf{activation}-maximization regularization. \cite{chen2019data} and \cite{haroush2020knowledge} compare the proposed data reconstruction methods with normal distribution and Gaussian distribution, respectively.
They show that the noise inputs hardly transfer useful knowledge. 

Addepalli~\etal~\cite{addepalli2020degan} propose a Data-enriching GAN (DeGAN) framework to retrieve representative samples from a trained classifier by using data from a related domain. Despite the generator, they introduce a proxy dataset and a discriminator to help data reconstruction. Following \cite{chen2019data} and \cite{ye2020data}, it adopts a \textbf{prediction} entropy and a diversity loss to penalize the generator. The proxy dataset contains related domain images with the original training data, such as new class data in class-incremental learning. The discriminator penalizes the generator to synthesize data impressions close to the proxy data. 

To tackle the problem of the diversity of generated samples, Han~\etal\cite{han2021robustness} address robustness and diversity in seeking data-free KD. They leverage an information entropy and a cross-entropy loss, and develop a diversity seeking regularization to prevent the generated samples from
being too similar with each other. Recently, Fang~\etal~\cite{fang2021contrastive} develop contrastive model inversion (CMI).
It models the data diversity as an optimizable objective to alleviate the mode collapse issue in data reconstruction. They adopt \textbf{BNS}, class prior (\textbf{cross-entropy}), and KL adversarial distillation to reconstruct data impressions. Moreover, they introduce another network as an instance discriminator upon the teacher to project features to a new embedding space. Then they conduct a positive view of the data impression $\tilde{x}$ by random augmentation and treat other instances as negative ones.
Upon this, they formulate a contrastive loss to promote the data diversity. 

Some generative reconstruction strategies~\cite{kariyappa2021maze,miura2021megex} are based on \textbf{gradient} estimation. But they are mainly for the model attack issue, where the model is regarded as a black box and only the predictions are accessible. We will detail these methods in Sec.~\ref{sec:attack}. 

In particular, instead of generating images from the teacher network with a series of prior, recently Chen~\etal~\cite{chen2021learning} propose to maximally utilize the massive available unlabeled data in the wild to address the data-free knowledge distillation problem. They first analyze the bound of distance between the outputs of the teacher and the student networks.
Then they collect data with noisy labels, which can be utilized to data-free noisy distillation (DFND). 
Similarly, Nayak~\etal~\cite{nayak2021effectiveness} investigate the effectiveness of arbitrary transfer sets in the data-free KD scenario, such as random noise and publicly available synthetic datasets. 
The experiments demonstrate that the arbitrary transfer sets is indeed effective in this scenario.

\begin{figure*}[!t]
\centering
\subfigure[Quantization]{
  \includegraphics[width=0.31\linewidth]{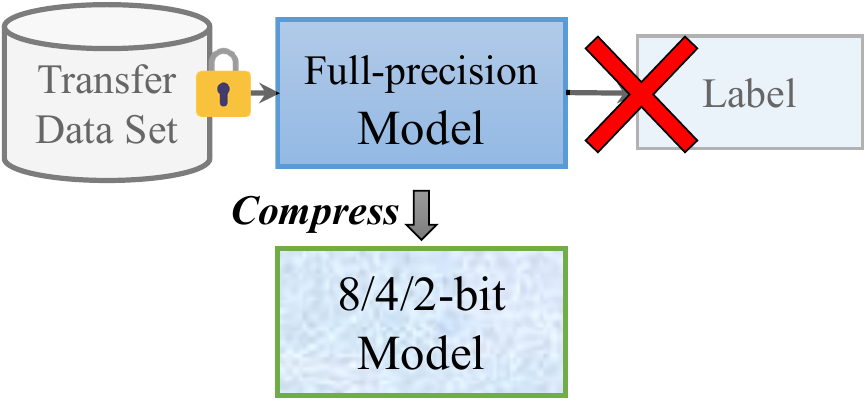}
}\hspace{5mm}
\subfigure[Incremental Learning]{
  \includegraphics[width=0.31\linewidth]{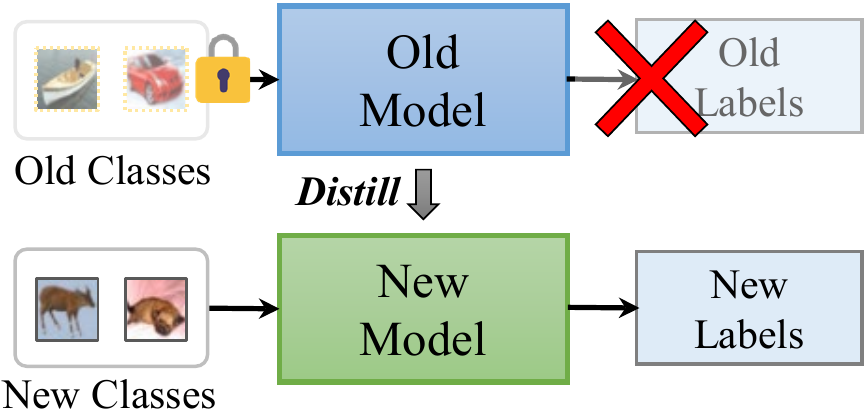}
}\hspace{5mm}
\subfigure[Model Attack]{
  \includegraphics[width=0.27\linewidth]{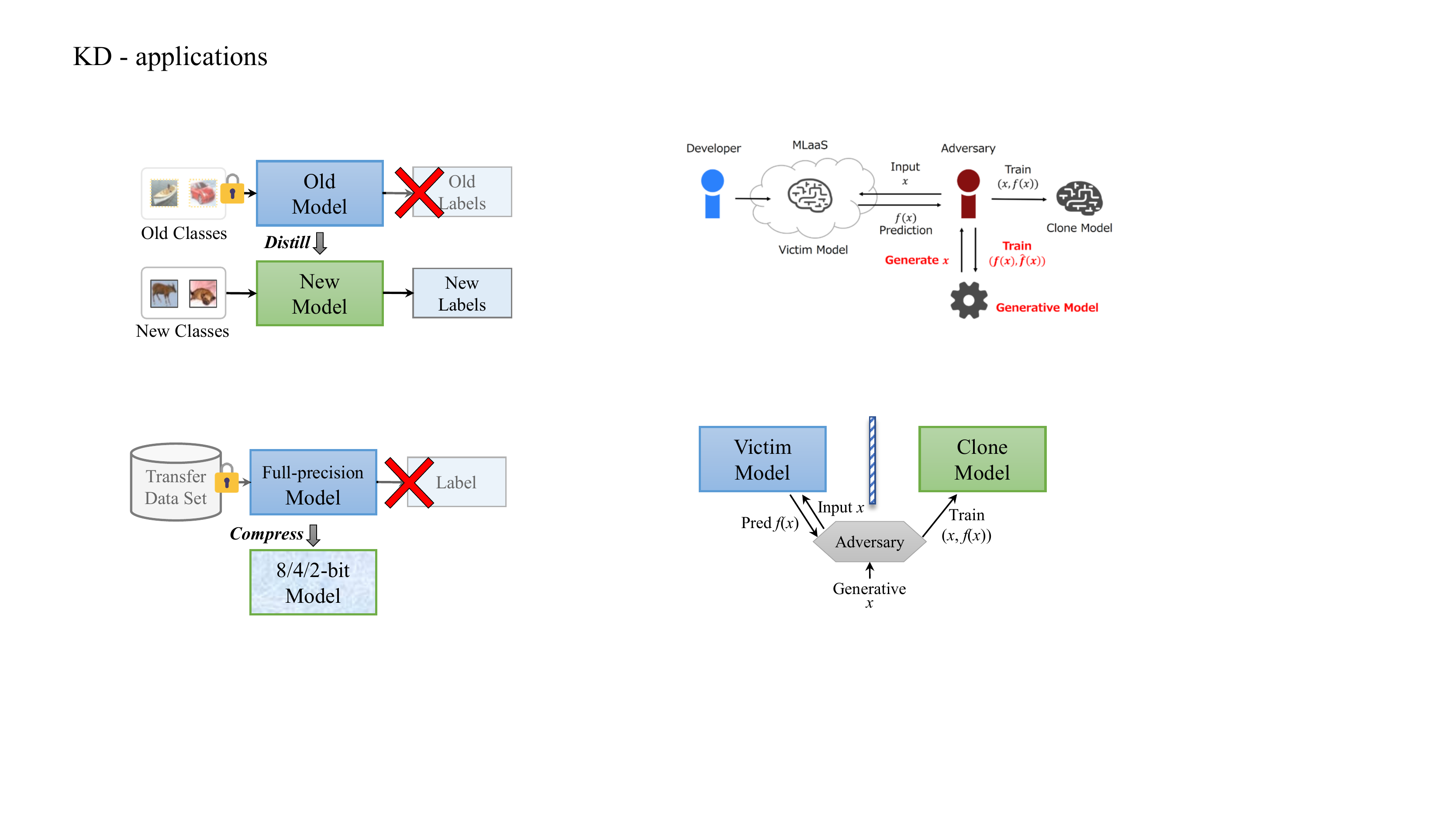}
}
\caption{Three major applications of data-free knowledge distillation. (a) Quantization: efficient model compression techniques for performing computations and storing tensors at lower bit-widths than floating point precision. (b) Incremental Learning: continual or lifelong learning, a branch of machine learning that involves processing incoming data from a data stream continuously. (c) Model Attack: an attack to violate intellectual property and privacy in which an adversary steals trained models in a cloud using only their predictions. }
\label{fig:app_dfkd}
\end{figure*}

\subsection{Adversarial Exploration}

As illustrated in the beginning of Sec.~\ref{sec:dfkd}, the core of DFKD is to obtain a reliable alternative dataset for knowledge distillation. The above two kinds of DFKD methods aim to reconstruct a set of realistic data impressions to replace the original training data in the distillation procedure. 
The quality of the data impressions determines the effect of knowledge transfer. However, there is still a big gap between the data impressions and the original data due to the unreliable and limited domain prior. 
Therefore, no matter how realistic the reconstructed data is, it is not a complete substitute for the original training data. 
To this end, some researchers~\cite{micaelli2019zero,fang2020data,liu2021zero} study an adversarial exploration framework to synthesize alternative data, as shown in Fig.~\ref{fig:dfkd}(c). 
Similar to generative reconstruction methods, a generator $\mathcal{G}$ is employed to synthesize samples, following Eqn.~\ref{eq:gen}. 
We would like to emphasize that the generator $\mathcal{G}$ is trained in a really adversarial procedure with the teacher-student architecture, which works as a joint discriminator to penalize $\mathcal{G}$. On the contrary, the learning objective of $\mathcal{G}$ is fixed in generative reconstruction because of the fixed teacher network $\mathcal{P}$. 
Hence, it is actually not a real adversarial learning paradigm, unable to have a learnable discriminator. 

In essence, the goal of DFKD is to craft a more lightweight student model $\mathcal{Q}$ by harnessing valuable knowledge from the well-trained teacher model $\mathcal{P}$.
And it does not require to access to real-world data.
To achieve this, the student $\mathcal{Q}$ can be updated by minimizing the model discrepancy:
\begin{equation}
  \mathcal{Q}^{*} = \arg\min_{\mathcal{Q}}\mathcal{D}(\mathcal{P}, \mathcal{Q}) \,, 
\end{equation}
where $\mathcal{D}$ is a model discrepancy metric. $\mathcal{D}(\mathcal{P}, \mathcal{Q})$ combines the teacher and the student into a joint discriminator. 
With the alternative data $\mathcal{G}(\boldsymbol{z})$ generated by $\mathcal{G}$, the model discrepancy can be formulated as 
\begin{equation}
  \mathcal{D}(\mathcal{P}, \mathcal{Q}; \mathcal{G}) = \mathbb{E}_{\boldsymbol{z}\sim p_z(\boldsymbol{z})}l(\mathcal{P}(\mathcal{G}(\boldsymbol{z})), \mathcal{Q}(\mathcal{G}(\boldsymbol{z}))) \,,
\end{equation}
in which $l(\cdot, \cdot)$ is a distance function, such as KL, MAE, MSE, \etc. 
The goal of the adversarial exploration phase is to encourage the generator $\mathcal{G}$ to synthesize more confusing training samples that lead a larger model discrepancy between the teacher and the student. 
Analogy to human learning process, the generator $\mathcal{G}$ is supposed to provide more advanced knowledge to help student solve more challenging problems. A straightforward way to achieve this goal is to update $\mathcal{G}$ (the student $\mathcal{Q}$ is fixed) with the negative discrepancy function, 
\begin{equation}
  \min_{\mathcal{G}} -\mathcal{D}(\mathcal{P}, \mathcal{Q}; \mathcal{G}) \,.
\end{equation}
In the knowledge transfer phase, the student $\mathcal{Q}$ is trained to mimic the teacher $\mathcal{P}$, with the generator $\mathcal{G}$ fixed:
\begin{equation}
  \min_\mathcal{Q} \mathcal{D}(\mathcal{P}, \mathcal{Q}; \mathcal{G}) \,.
\end{equation}
The adversarial exploration and knowledge transfer stages alternate in each round of training. From an adversarial learning perspective, $\mathcal{G}$ aims to maximize the model discrepancy, while $\mathcal{Q}$ is supposed to minimize it. Different from noise optimization or generative reconstruction, the adversarial exploration phase does not focus on generating realistic and human-recognizable samples, but works on synthesizing hard samples that are more worth learning. 
Analogous to the knowledge form used in traditional KD~\cite{wang2021knowledge,gou2021knowledge}, the model discrepancy $\mathcal{D}$ can be calculated with the outputs or intermediate features of the two models. Therefore, according to the discrepancy calculation, the adversarial exploration-based DFKD methods can be divided into output-based and feature-based categories. 

ZSKT~\cite{micaelli2019zero} achieves data-free knowledge transfer by firstly training an adversarial generator to search for images on which the student poorly matches the teacher, and then using them to train the student. It adopts the KL divergence as a model discrepancy metric and introduces attention transfer~\cite{zagoruyko2016paying} to enhance the knowledge distillation phase. To avoid producing decayed gradients on converged samples, DFAD~\cite{fang2020data} advocates to utilize Mean Absolute Error (MAE) as a model dependency function. Moreover, DFAD is the first DFKD approach that can be applied to semantic segmentation, due to the general and model-agnostic discrepancy measures. It firstly claims that using realistic samples are not the only way for knowledge distillation, which has a profound impact on DFKD. Recently, FastDFKD~\cite{fang2022up} introduces an efficacious scheme with a meta generator to accelerate the DFKD process, achieving a 100$\times$ faster speed. 

Based on DFAD~\cite{fang2020data}, Zhao~\etal{} develop a novel dual discriminator adversarial distillation (DDAD)~\cite{zhao2021dual} framework and conduct more comparison experiments on classification and segmentation tasks. 
To solve an inherent biased sample generation problem, Lee~\etal~\cite{lee2021zero} propose the C-ZSKD algorithm to increase the variance of the adversarial sample distribution by using the convolution of probability distributions and Taylor series approximation. AVKD~\cite{tang2021adversarial} extends the ZSKT~\cite{micaelli2019zero} and formulates the adversarial exploration process as variational autoencoders (VAE). To prevent catastrophic forgetting in DFKD, Binici~\etal~\cite{binici2021preventing} propose to maintain a dynamic collection of synthetic samples over time. And, to mitigate the distribution mismatch between the synthetic and real data, they exploit the three regularization losses in DAFL~\cite{chen2019data} and the discrepancy maximization strategy in DFAD~\cite{fang2020data}.

\begin{table*}[!t]
  \centering
  \caption{Applications of DFKD on Quantization and Pruning. The middle three columns are for the selection of data generation algorithms. `Prior' is for the types of prior regularization, abbreviated as their initials: \textbf{A}ctivation, \textbf{B}NS, \textbf{D}iscrepancy, \textbf{G}radient, and \textbf{P}rediction (the same below). \cmark / \xmark~ is for yes / no. `\textbf{Q}uantization or \textbf{P}runing' is for the compression approach. `Only Classification' is for checking whether the method is only for classification task. `Diversity' is for checking whether the method considers the diversity of synthesized samples. }
  \resizebox*{\linewidth}{!}{
    \begin{tabular}{c|c|c|ccc|c|c|c|c}
    \toprule
    Method & Publication & Year  & \multicolumn{1}{c}{\makecell[c]{Noise\\ Optimization}} & \multicolumn{1}{c}{\makecell[c]{Generative\\ Reconstruction}} & \multicolumn{1}{c|}{\makecell[c]{Adversarial\\ Exploration}} & Prior & \multicolumn{1}{c|}{\makecell[c]{\textbf{Q}uantization\\ or \textbf{P}runing}} & Only Classification & Diversity \\
    \midrule
    ZeroQ~\cite{cai2020zeroq} & CVPR  & 2020  & \cmark   &       &       & B & Q     & \xmark    & \xmark \\
    GDFQ~\cite{xu2020generative}  & ECCV  & 2020  &       & \cmark   &       & B, P  & Q     & \cmark   & \xmark \\
    Choi~\etal~\cite{choi2020data} & CVPRW & 2020  &       & \cmark   & \cmark   & B, D, P & Q     & \cmark   & \xmark \\
    Zhu~\etal~\cite{zhu2020towards} & arXiv & 2020  &       &       & \cmark   & D & Q+P   & \cmark   & \xmark \\
    Horton~\etal~\cite{horton2020layer} & arXiv & 2021  &       & \cmark   &       & B & Q+P   & \cmark   & \xmark \\
    ZAQ~\cite{liu2021zero}   & CVPR  & 2021  &       & \cmark   & \cmark   & A, D  & Q     & \xmark    & \xmark \\
    DSG~\cite{zhang2021diversifying,qin2021diverse}   & CVPR  & 2021  & \cmark   &       &       & B & Q     & \cmark   & \cmark \\
    He~\etal~\cite{he2021generative} & CVPRW & 2021  &       & \cmark   &       & B, P  & Q     & \cmark   & \xmark \\
    MixMix~\cite{li2021learning} & ICCV & 2021 & \cmark  &       &       & B, P, patch   & Q+P   & \cmark   & \xmark \\
    FDDA~\cite{zhong2021fine}  & arXiv & 2021  &       & \cmark   &       & B, P  & Q     & \cmark   & \xmark \\
    IntraQ~\cite{zhong2021IntraQ} & arXiv & 2021  & \cmark   &       &       & B, patch & Q     & \cmark   & \cmark \\
    \bottomrule
    \end{tabular}
  }
  \label{tab:app_qp}
\end{table*}

\begin{table*}[!t]
  \centering
  \caption{Applications of DFKD on Model Attack. `Black-box' is for checking whether black-box attack. `Distributed' is for checking whether the models are in distributed learning. }
    \begin{tabular}{c|c|c|ccc|c|c|c}
    \toprule
    Method & Publication & Year  & \multicolumn{1}{c}{\makecell[c]{Noise\\ Optimization}} & \multicolumn{1}{c}{\makecell[c]{Generative\\ Reconstruction}} & \multicolumn{1}{c|}{\makecell[c]{Adversarial\\ Exploration}} & Prior & Black-box & Distributed \\
    \midrule
    DLG~\cite{zhu2019deep}   & NeurIPS & 2019  & \cmark &       &       & G     & \xmark    & \cmark \\
    Geiping~\etal~\cite{jonas2020inverting} & NeurIPS & 2020  & \cmark &       &       & G     & \xmark    & \cmark \\
    GradInversion~\cite{yin2021see} & CVPR  & 2021  & \cmark &       &       & B, G, P & \xmark    & \cmark \\
    DaST~\cite{zhou2020dast} & CVPR & 2020 &      &       & \cmark & D, P & \cmark & \xmark \\
    MAZE~\cite{kariyappa2021maze}  & CVPR  & 2021  &       &       & \cmark   & D, G, P  & \cmark & \xmark \\
    DFME~\cite{truong2021data}  & CVPR  & 2021  &       &       & \cmark   & D, G, P  & \cmark & \xmark \\
    Wang~\etal~\cite{wang2021delving} & CVPR  & 2021  &       & \cmark   &       & P & \cmark & \xmark \\
    CDI~\cite{dong2021deep}   & arXiv & 2021  & \cmark   &       &       & B, P  & \xmark & \xmark \\
    MEGEX~\cite{miura2021megex} & arXiv & 2021  &       &       & \cmark   & D, G, P  & \cmark & \xmark \\
    ZSDB3KD~\cite{wang2021zero} & ICML  & 2021  & \cmark   &       &       & P, boundary & \cmark & \xmark \\
    Yue~\etal~\cite{yue2021black} & RecSys & 2021  & \cmark   &       &       & P & \cmark & \xmark \\
    FE-DaST~\cite{yu2021fe} & Computers \& Security & 2021 &       & \cmark  & \cmark   & D, P & \cmark & \xmark \\ 
    \bottomrule
    \end{tabular}
  \label{tab:app_attack}
\end{table*}

\subsection{Application}
\label{sec:app1}

Over the last few years, a variety of data-free knowledge distillation methods have been widely used to solve the problem of model compression and transfer learning in different intelligent applications. In this section, we mainly review the applications of DFKD on quantization \& pruning, incremental learning, and model attack, as shown in Fig.~\ref{fig:app_dfkd}. Not limited to computer vision, there are also some applications in NLP and graph neural networks (GNN). 

\subsubsection{Quantization \& Pruning}

Quantization and pruning are two major promising approaches for deep model compression. Quantization aims to store parameters with fewer bits, so that computation can be executed on integer-arithmetic units rather than on power-hungry floating-point ones. And pruning focuses on removing the redundant convolutional filters or linear connections in deep neural networks to reduce storage and computation. Due to the performance degradation of quantized or pruned models, it is necessary to retrain them on the original training data, even with the guide of pre-trained models. In data-free scenarios, some DFKD frameworks can be applied to model quantization (or pruning), where the teacher is the pre-trained full-precision model, and the student is the quantized (or pruned) model, as shown in Fig.~\ref{fig:app_dfkd}(a). We list some typical applications of DFKD on quantization and pruning in Tab.~\ref{tab:app_qp}. 

ZeroQ~\cite{cai2020zeroq} achieves zero-shot/data-free post-quantization by reconstructing data impressions via BNS, and supports mix-precision quantization with a Pareto frontier based determination. On classification and object detection tasks, it is verified to outperform previous post-quantization methods without fine-tuning~\cite{nagel2019data,li2019dac}. The noise optimization only constraints the data impressions to match the BN statistics, causing overfitting and limited diversity of synthesized samples. Thus, some works~\cite{zhang2021diversifying,qin2021diverse,zhong2021IntraQ} focus on increasing the diversity of data impressions in noise optimization to expand the coverage space of synthetic samples. Specifically, DSG~\cite{zhang2021diversifying,qin2021diverse} enhances the diversity of data by two techniques: (1) slacking the alignment of the feature statistics in BN layers to relax the distribution constraint; applying the layer-wise enhancement to reinforce specific layers for different data samples. IntroQ~\cite{zhong2021IntraQ} randomly crops a local patch for object reinforcement, which can retain the intra-class heterogeneity of synthetic samples. It conducts experiments both on quantization and pruning. Inspired by generative DFKD, most data-free quantization/pruning methods~\cite{xu2020generative,choi2020data,horton2020layer,he2021generative,zhong2021fine} exploit a generator for more efficient data synthesis and knowledge transfer, avoiding replicating the noise optimization process. They utilize different knowledge prior based regularization, the primary of which is BNS. 
In particular, Choi~\etal~\cite{choi2020data} combine generative reconstruction and adversarial exploration to synthesize discriminative and recognizable images. Our ZAQ~\cite{liu2021zero} mainly works by exploring and transferring information of individual samples and their correlations, mitigating the gap between the full-precision and quantized models. Inspired by \cite{ye2020data}, Horton~\etal~\cite{horton2020layer} brake the problem of data-free network compression into a number of independent layer-wise compression.
They introduce layer-wise data generation and cross-layer equalization to approximate the compressed model. 
From Tab.~\ref{tab:app_qp}, we can conclude that most of the data-free compression methods are limited to classification tasks.
And only ZeroQ~\cite{cai2020zeroq} and our ZAQ~\cite{liu2021zero} can work on detection or segmentation. 

In fact, most of the above methods can be regarded as general DFKD frameworks used for learning quantized or pruned student models.
This is because the data reconstruction and knowledge transfer processes are generalized. There needs more efforts on specific data-free algorithms for quantization and pruning. For instance, the mix-precision quantization and filter pruning strategies are supposed to be determined under data-free scenarios. In addition to promoting the diversity of synthetic samples, how to take advantages of the intermediate feature maps for data-free quantization or pruning is also worthy to be further studied. 

\subsubsection{Incremental Learning}

\begin{table*}[!t]
  \centering
  \caption{Applications of DFKD on other tasks. `Distillation loss' is for the distillation loss function used in knowledge transfer phase: $\mathcal{L}_{PT}$ means the patient knowledge distillation loss~\cite{sun2019patient} for BERT model, $\mathcal{L}_{Det}$ represents the object detection loss, and $\mathcal{L}_{AT}$ is the attention transfer loss. }
    \begin{tabular}{c|c|c|c|ccc|c|c}
    \toprule
    Method & Task  & Publication & Year  & \multicolumn{1}{c}{\makecell[c]{Noise \\ Optimization}} & \multicolumn{1}{c}{\makecell[c]{Generative\\ Reconstruction}} & \multicolumn{1}{c|}{\makecell[c]{Adversarial\\ Exploration}} & Prior & Distillation loss \\
    \midrule
    AS-DFD~\cite{ma2020adversarial} & Text Classification & EMNLP & 2020  & \cmark   &       &       & P, mask & $\mathcal{L}_{KL} + \mathcal{L}_{PT}$ \\
    Rashid~\etal\cite{rashid2020towards} & Text Classification & EMNLP & 2021 &      &       & \cmark & D  & $\mathcal{L}_{KL}$ \\
    Bhogale\etal~\cite{bhogale2020data} & Semantic Segmentation & arXiv & 2020  &       & \cmark   &       & P, diversity & $\mathcal{L}_{KL}$ \\
    GFKD~\cite{deng2021graph}  & Graph Classification & IJCAI & 2021  & \cmark   &       &       & B, P  & $\mathcal{L}_{KL}$ \\
    Zhang~\etal~\cite{zhang2021data} & Super-Resolution & CVPR  & 2021  &       &       & \cmark   & D     & $\mathcal{L}_{1}$ \\
    Chawla~\etal~\cite{chawla2021data} & Object Detection & WACV  & 2021  & \cmark   &       &       & B     & $\mathcal{L}_{2}$ \\
    FEDGEN~\cite{zhu2021data} & Federated Learning & ICML  & 2021  &       & \cmark   &       & P     & $\mathcal{L}_{CE} + \mathcal{L}_{KL}$ \\
    DFED~\cite{hao2021datafree}  & Image Classification & ACM MM & 2021  &       & \cmark   & \cmark   & B, D, P & $\mathcal{L}_{KL}$ \\
    Nayak~\etal~\cite{nayak2021classification} & Object Detection & BMVC  & 2021  & \cmark   &       &       & P, diversity & $ \mathcal{L}_{Det} + \mathcal{L}_{2}$ \\
    MosaicKD~\cite{fang2021mosaicking} & Out-of-domain KD & NeurIPS & 2021  &       &       & \cmark   & D, P  & $\mathcal{L}_{KL}$ \\
    DFPU~\cite{tang2021datafree} & PU Leaning & ICONIP & 2021 &    & \cmark   &   & A, P & $\mathcal{L}_{KL} + \mathcal{L}_{AT}$ \\ 
    \bottomrule
    \end{tabular}
  \label{tab:app_others}
\end{table*}

Incremental learning (also known as a subset of continual or lifelong learning) is a learning paradigm that enables a model to continuously update with new incoming data, while avoiding storing the old data and training the model repeatedly~\cite{parisi2019continual}. A straightforward approach is to fine-tune a pre-trained model on new data, but this could damage the performance on the old task, due to the catastrophic forgetting problem. So replaying the old data or distilling the old knowledge is necessary to prevent catastrophic forgetting. From a perspective of data-free knowledge transfer, this issue can be addressed by reconstructing some alternative data to distill the old knowledge, without accessing any previous data (Fig.~\ref{fig:app_dfkd}(b)). To this end, Choi~\etal~\cite{choi2021dual} introduce a dual-teacher information distillation framework to synthesize old data and feed to the current model trained on the new data. They exploit the prediction cross-entropy and BNS losses as prior of old data generation. As for the distribution process, they utilize a variational information distillation and a cosine-similarity constraint to transfer and reserve the old knowledge. 
Following DAFL~\cite{chen2019data} and DeepInversion~\cite{yin2020dreaming}, Smith~\etal~\cite{smith2021always} employ a combination loss of information entropy, class cross-entropy, smoothness prior, and BNS alignment to synthesize images. In addition, to prevent fails for common class-incremental benchmarks, they propose a modified cross-entropy training and importance-weighted feature distillation. Recently, Huang~\etal~\cite{huang2021half} propose a half-real half-fake distillation framework for class-incremental semantic segmentation.
In the framework, the fake data is generated with the pixel-level cross-entropy loss and image regularization, and the knowledge preservation is achieved by standard segmentation distillation~\cite{liu2019structured}.

\subsubsection{Model Attack}
\label{sec:attack}

Model extraction attacks, or model substitute attacks, allow an adversary to clone or steal the functionality of a black-box machine learning model, compromising its confidentiality. To query the target model and steal information from the prediction in absence of real data, some works inherit the idea of DFKD as shown in Fig.~\ref{fig:app_dfkd}(c). 
Compared with the settings of standard DFKD, the only difference for the model extraction attack is that the intermediate feature maps or gradients of the model are inaccessible. Tab.~\ref{tab:app_attack} lists some typical applications of DFKD on model attacks. In distributed learning scenarios, some works~\cite{zhu2019deep,jonas2020inverting,yin2021see} attempt to recover the real data with the shared gradients from other training nodes. In this way, the attacked model works as a white box. 
Inspired by DeepInversion~\cite{yin2020dreaming}, GradInversion~\cite{yin2021see} innovatively converts random noise into natural images by matching gradients while regularizing image fidelity through BNS and cross-entropy priors. DaST~\cite{zhou2020dast}, MAZE~\cite{kariyappa2021maze}, DFME~\cite{truong2021data}, MEGEX~\cite{miura2021megex}, and FE-DaST~\cite{yu2021fe} are the attack methods that exploit the model discrepancy in data generation and knowledge transfer. 
To enable a highly accurate model extraction attack, MAZE~\cite{kariyappa2021maze}, DFME~\cite{truong2021data}, and MEGEX~\cite{miura2021megex} all utilize a zeroth-order gradient estimation to perform optimization in the black-box setting. Following DAFL~\cite{chen2019data}, \cite{wang2021zero} and \cite{yu2021fe} introduce the prediction constraints of a victim model for generative reconstruction. In particular, ZSDB3KD~\cite{wang2021zero} proposes concept of decision-based black-box knowledge distillation for the first time. CDI~\cite{dong2021deep} intends to optimize random noises into data impressions with BNS prior, which is actually unavailable in black-box setting. Yue~\etal~\cite{yue2021black} propose an API-based model extraction method for sequential recommender systems via limited-budget synthetic data generation and knowledge distillation. With the surrogate data, they investigate two downstream attacks, \ie, profile pollution attacks and data poisoning attacks. 

\subsubsection{Other Tasks}

\textbf{Other CV Tasks.} \cite{chawla2021data} and \cite{nayak2021classification} are two works on object detection.
\cite{chawla2021data} designs an automated bounding box and category sampling scheme based on DeepInversion~\cite{yin2020dreaming}. \cite{nayak2021classification} utilizes object/patch-level impression and diversity to enhance the data generation. Different from image classification, each segmentation image contains multiple categories of pixels, which requires the data impression also represents diverse classes. To this end, Bhogale~\etal~\cite{bhogale2020data} make use of the DeGAN~\cite{addepalli2020degan} training framework and propose a novel loss function to enforce the diversity of semantic image synthesis, for segmentation task. Zhang~\etal~\cite{zhang2021data} follow DFAD~\cite{fang2020data} and present a progressive training algorithm for super-resolution in data-free setting. FEDGEN~\cite{zhu2021data} is an ensemble DFKD approach in heterogeneous federated learning or decentralized machine-learning situations. Moreover, DFKD can be exploited to tackle the out-of-domain distillation task~\cite{fang2021mosaicking}, which allows us to conduct KD using only out-of-domain data that can be readily obtained at a very low cost. 

\textbf{Other non-CV Tasks.} In addition to the above visual applications, DFKD has been widely applied to other fields, including but not limited to NLP, GNN, and federated learning, as shown in Fig.~\ref{tab:app_others}. 
Different from continuous and real-valued image data, the vocabulary-based representations of texts are discrete in NLP models.
As such, they cannot be directly generated by the common algorithms used in vision models. To tackle the challenge of DFKD for transformer-based NLP models (\eg, BERT~\cite{devlin2018bert}), AS-DFD~\cite{ma2020adversarial} introduces a plug and play embedding guessing algorithm and an adversarial self-supervised module to search realistic pseudo embeddings in the first stage. Similar to \cite{bhardwaj2019dream,nayak2019zero,yin2020dreaming}, the obtained alternative embeddings in the first stage are utilized for knowledge distillation in the second stage, which follows PKD~\cite{sun2019patient}. Rashid~\etal~\cite{rashid2020towards} extend the ZSKT~\cite{micaelli2019zero} to NLP models and introduce out-of-domain data to assist the text classification tasks. GFKD~\cite{deng2021graph} is the first dedicated DFKD method designed for GNN on graph classification task. It extends the priors commonly used in convolutional neural networks to GNN, and develops a gradient estimator for graph generation.

\subsection{Discussion}

We have illustrated three kinds of DFKD frameworks and their applications in the above. In this section, we briefly discuss the advantages and challenges of each kind of DFKD methods. 

The noise optimization method can directly convert a random noise to a realistic natural image, which is recognizable and explainable for data-free knowledge distillation. 
It is worth noting that the quality of synthetic data impressions depends on the selection of knowledge priors. 
For example, the activation records only support simple and low-resolution image generation, while multi-layer BNS can provide accurate domain constraints for real data and becomes the most popular information source for data reconstruction. 
The cross entropy or information entropy of the class prediction plays a decisive role in the synthesis of identifiable samples, but it is limited to classification task. 
Moreover, the noise optimization has an independent two-stage training scheme, \ie, data reconstruction and knowledge distillation. The effectiveness of the knowledge distillation depends on the quality of reconstructed data. 
However, each reconstructed data impression needs to go through hundreds or even thousands of iterative optimization, which is extremely time-consuming. 
Before sufficient data impressions are obtained, we cannot perform reliable knowledge distillation.
As such, it usually takes plenty of time to update the parameters for noise optimization. 

Generative reconstruction methods can overcome the weakness of the noise optimization methods by employing a generator for data synthesis. Though facing the challenge of pattern collapse, it reduces the training time by parameterizing the distribution of fake data during iterative generation.
Nevertheless, due to the limited capability of generators, this kind of method barely supports high-resolution image generation. 
By contrast, with the BNS prior, the noise optimization methods can synthesize extremely realistic and high-resolution images. 

To get rid of the constraints of recognition and visual content of alternative images, adversarial exploration methods achieve DFKD via a real adversarial learning framework. 
Different from generative reconstruction framework, a joint teacher-student discriminator is updated synchronously with the data generator.
Thus, it can pay attention to exploring more discriminative, more diverse, and harder samples, rather than recognizable data impressions. 
In addition, it is task agnostic and model agnostic, not only supporting classification tasks. 
The challenge in this kind of methods is that the generated samples are not entirely reliable.
Therefore, some unrelated samples could be barely effective in knowledge transfer.

\section{Source Data-Free Domain Adaptation}
\label{sec:sfda}

The goal of domain adaptation is to reuse the source domain knowledge for the target domain via a source data-driven two-stage framework. Firstly, source data is collected and annotated to train the model $\mathcal{P}$. Secondly, the adaptation phase is to fine-tune the pre-trained model $\mathcal{P}$ with unlabeled target data and labeled source data, obtaining the target model $\mathcal{Q}$. 
In the source data-free domain adaptation (SFDA) scenario, only the unlabeled target data $X^t$ can be utilized to distill valuable knowledge from the pre-trained source model $\mathcal{P}$. To achieve this, one approach is to take advantages of the target samples via self-supervised training. Another approach is to reconstruct virtual source data for knowledge transfer. In this section, we denote the source and target models as $\mathcal{P}$ and $\mathcal{Q}$, respectively.
Both of them usually share the same model architecture without specific instruction.

\subsection{Self-supervised Training}

\begin{figure}[!t]
\centering
\subfigure[Clustering]{
  \includegraphics[width=0.7\linewidth]{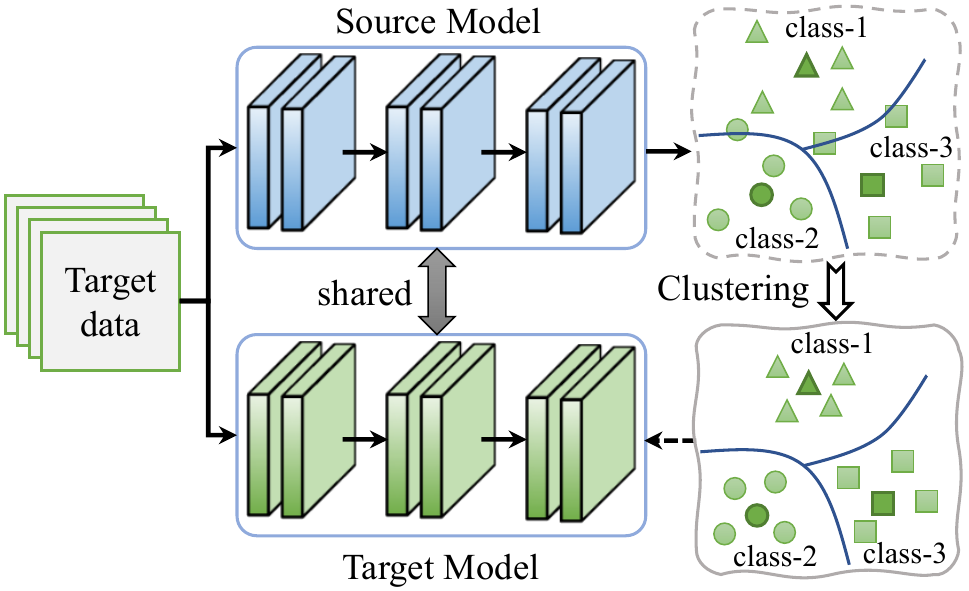}
}
\subfigure[Filtering]{
  \includegraphics[width=0.7\linewidth]{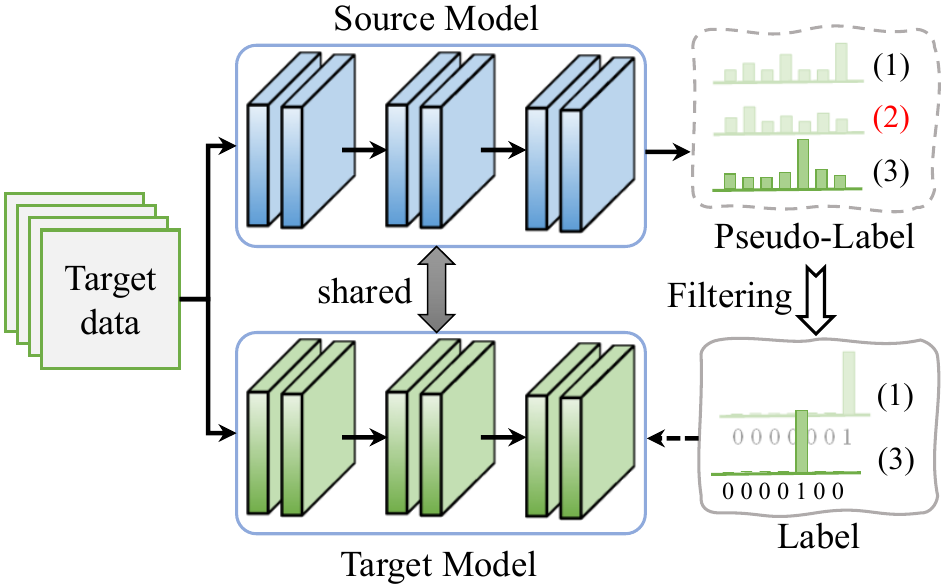}
}
\caption{SFDA with Self-supervised Training: (a) pseudo-label clustering; (b) pseudo-label filtering. The source and target models share the same architecture in the two frameworks. The different symbols in (a) represent samples belonging to different classes. }
\label{fig:sfda1}
\end{figure}

\cite{chidlovskii2016domain} and \cite{nelakurthi2018source} are two pioneering studies of SFDA, but they mainly study to tackle the problem with few labeled target data. Only with the pre-trained model $\mathcal{P}$ and the unlabeled target dataset $X^t$, a straightforward adaptation way is to generate a set of noisy pseudo labels,
\begin{equation}
  \label{eq:pseudo}
  \tilde{Y}^t = \arg\max \mathcal{P}(X^t) \,,
\end{equation}
and then fine-tune the target model $\mathcal{Q}$ by 
\begin{equation}
  \min_{\mathcal{Q}} \mathcal{L}_{\text{adap}}(\mathcal{Q} | X^t, \tilde{Y}^t) \,,
\end{equation}
where $\mathcal{L}_{\text{adap}}$ is the adaptation loss function. 
Indeed, this could misguide the target model due to the existing error-prone pseudo labels. 

Originated from semi-supervised learning, some source-driven DA methods~\cite{vu2019advent,chen2019maxsquare} iteratively refine the model by exploring the most confident pseudo labels generated for the target set. In this scenario, the source set is involved to joint training, avoiding catastrophic forgetting of source knowledge. 
In the absence of source data, the target data with reliable pseudo labels $\tilde{Y}^t$ is critical to calibrate the model $\mathcal{Q}$ in turbulent self-supervised target learning. There are two major manners to explore the reliable and valuable target samples for self-supervision, \ie, pseudo-label clustering and filtering, as shown in Fig.~\ref{fig:sfda1}.

\subsubsection{Pseudo-label Clustering}

As depicted in Fig.~\ref{fig:sfda1}(a), the noisy pseudo labels could be firstly generated by the source model with target data input.
Then they could be further categorized through clustering algorithms. 
Based on the target data with calibrated pseudo labels, the target model could be learned. 
The intuition of pseudo-label clustering is that pre-trained source models can partially work on related target domains by predicting correctly on part of target samples. 

As an early study, Liang~\etal~\cite{liang2019distant} build a method upon the nearest centroid classifier, seeking a subspace of target centroid shift and iteratively minimizing the shift. Later, inspired by a transfer learning setting known as Hypothesis Transfer Learning (HTL)~\cite{kuzborskij2013stability}, Liang~\etal{} propose the Source HypOthesis Transfer (SHOT)~\cite{liang2020we} framework.
It freezes the classifier (hypothesis) of the source model and learns the target-specific feature extractor.
This is realized by exploiting both information maximization and self-supervised pseudo labeling to implicitly align representations from the target domains to the source hypothesis. 
As a matter of fact, the information maximization is to reduce the uncertainty of predictions iteratively, which is similar to the data-driven DA.
Moreover, the pseudo labels are obtained by weighted k-means clustering~\cite{caron2018deep} and nearest centroid classifier. 
They further extend SHOT with a new labeling transfer strategy, dubbed as SHOT++~\cite{liang2021source}.
It separates the target data into two splits based on the confidence of predictions, and then employs semi-supervised learning to improve the accuracy of less-confident predictions in the target domain. In addition, Tang~\etal~\cite{tang2021model} also follow SHOT~\cite{liang2020we}, and introduce the gradual KD and Mixup~\cite{zhang2017mixup} techniques. Chen~\etal~\cite{chen2021self} utilize adaptive batch normalization to update the target data and then generate pseudo labels via deep transfer clustering~\cite{han2019learning}. In addition, they split the target set into clean and noisy part for stable training with exponential momentum average (EMA). G-SFDA~\cite{yang2021generalized} is a generalized SDFA method that unifies local structure clustering and sparse domain attention. The former utilizes local neighbor information in feature space to cluster each target example with its nearest neighbors for self-supervised adaptation, while the latter aims to keep information of the source domain through regularizing the gradient of back propagation. Tang~\etal~\cite{tang2021nearest} propose a semantic constraint hidden in the local geometry of individual data to encourage robust clustering on the target domain.
They additionally introduce semantic credibility constraints in the new geometry. 
Yang~\etal~\cite{yang2021exploiting} exploit the intrinsic neighborhood structure by defining local affinity of the target data, and encourage class consistency with neighborhood affinity. 

Apart from tackling the basic SFDA challenges, there are some works focusing on other specific SFDA scenarios. USFDA~\cite{kundu2020universal} is a two-stage learning framework to address universal source-free domain adaptation. In the procurement stage, it equips the model for future source-free deployment with an artificially generated negative dataset. This encourages well-separated tight source clusters. In the deployment stage, it defines a source-free adaptation objective by utilizing a source similarity metric. To solve the problem of multiple multi-source data-free adaptation, Ahmed~\etal~\cite{ahmed2021unsupervised} deploy information maximization and clustering-based~\cite{caron2018deep} pseudo-label strategy on the weighted combination of target soft labels from all the source models. Based on SHOT\cite{liang2020we}, Agarwal~\etal~\cite{agarwal2021unsupervised} pose a new problem of unsupervised robust domain adaptation in the source-free setting. Analogously, Yang~\etal~\cite{yang2021transformer} extend SHOT~\cite{liang2020we} to the Transformer-based networks and present a self-distillation SFDA framework with EMA to transfer clustered label knowledge to the target network. Unlike existing methods that focus on the training phase, T3A~\cite{iwasawa2021test} concentrates on the test phase in domain generalization. It computes a pseudo-prototype representation for each class with unlabeled target data, and then classify each sample based on its distance to the pseudo prototypes.

\subsubsection{Pseudo-label Filtering}
Despite the absence of source data, some target samples could spread around the corresponding source prototypes, and are very similar to the source domain. Therefore, these target samples could be used to approximate the source domain. This kind of SFDA methods usually filters the target pseudo-labels by splitting the target data into two subsets, \ie, pseudo-source set and remaining target set.
They correspond to source hypothesis keeping and target knowledge exploration, respectively. 
Fig.~\ref{fig:sfda1}(b) depicts the principle of the pseudo-label filtering strategy. 

Kim~\etal~\cite{kim2020domain} observe that target samples with a lower self-entropy measured by the pre-trained source model are more likely to be classified correctly. To this end, they select reliable samples with the entropy values less than 0.2 for target model training.
The pseudo labels of the selected samples are generated by the source model. 
BAIT~\cite{yang2020unsupervised} splits the current batch into two sets.
It finds potentially wandering features for minimizing the disagreement across the dual-classifier architecture. 
CAiDA~\cite{dong2021confident} is a pioneering exploration of knowledge adaptation from multiple source domains to the unlabeled target domain in the absence of source data. 
It develops a semantic-nearest confident anchor to select pseudo labels for self-supervised adaptation.
And it devises a class-relationship aware consistency loss to ensure the semantic consistency. 
Moreover, it theoretically proves that multiple source models could improve the possibility of obtaining more reliable pseudo labels under some mild assumptions. 
Instead of splitting the target set rigidly, Huang~\etal~\cite{huang2021model} build a historical contrastive learning (HCL) paradigm and allocate specific weight to each sample to decide its influence in self-training. Concretely, they design two historical contrastive discrimination strategies from instance level and category level, which exploit historical source hypothesis and learn discriminative target representations. $\text{A}^2\text{Net}$~\cite{xia2021adaptive} is an adaptive adversarial network including three components: (1) The adaptive adversarial inference is to discover source-similar and source-dissimilar target samples with a dual-classifier architecture.
(2) The contrastive category-wise matching is for category alignment between the two parts of target samples.
(3) The self-supervised rotation aims to enhance the model to learn additional semantics. 
Du~\etal~\cite{du2021generation} also split the target data into two disjoint parts, \ie, the pseudo-source part and the remaining target part, and adopt the Mixup~\cite{zhang2017mixup} strategy to align the distributions. 

Different from common SFDA scenarios, Zhang~\etal~\cite{zhang2021unsupervised} study the setting that the pre-trained source model is treated as a black box and only the predictions are accessible. 
They propose an iterative noisy label learning (IterNLL) algorithm which alternates between improving the target model with a filtered target subset and updating the noisy labels by a uniform prior assumption. 

Notably, the SFDA methods with pseudo-label clustering only naturally support classification tasks due to the limitation of clustering algorithms. But the pseudo-label filtering algorithm can be applied to other complex tasks, \eg, semantic segmentation and object detection, because of the divisibility of the prediction maps.

\subsection{Virtual Source Knowledge Transfer}

\begin{figure}[!t]
\centering
\subfigure[Source Impression]{
  \includegraphics[width=0.7\linewidth]{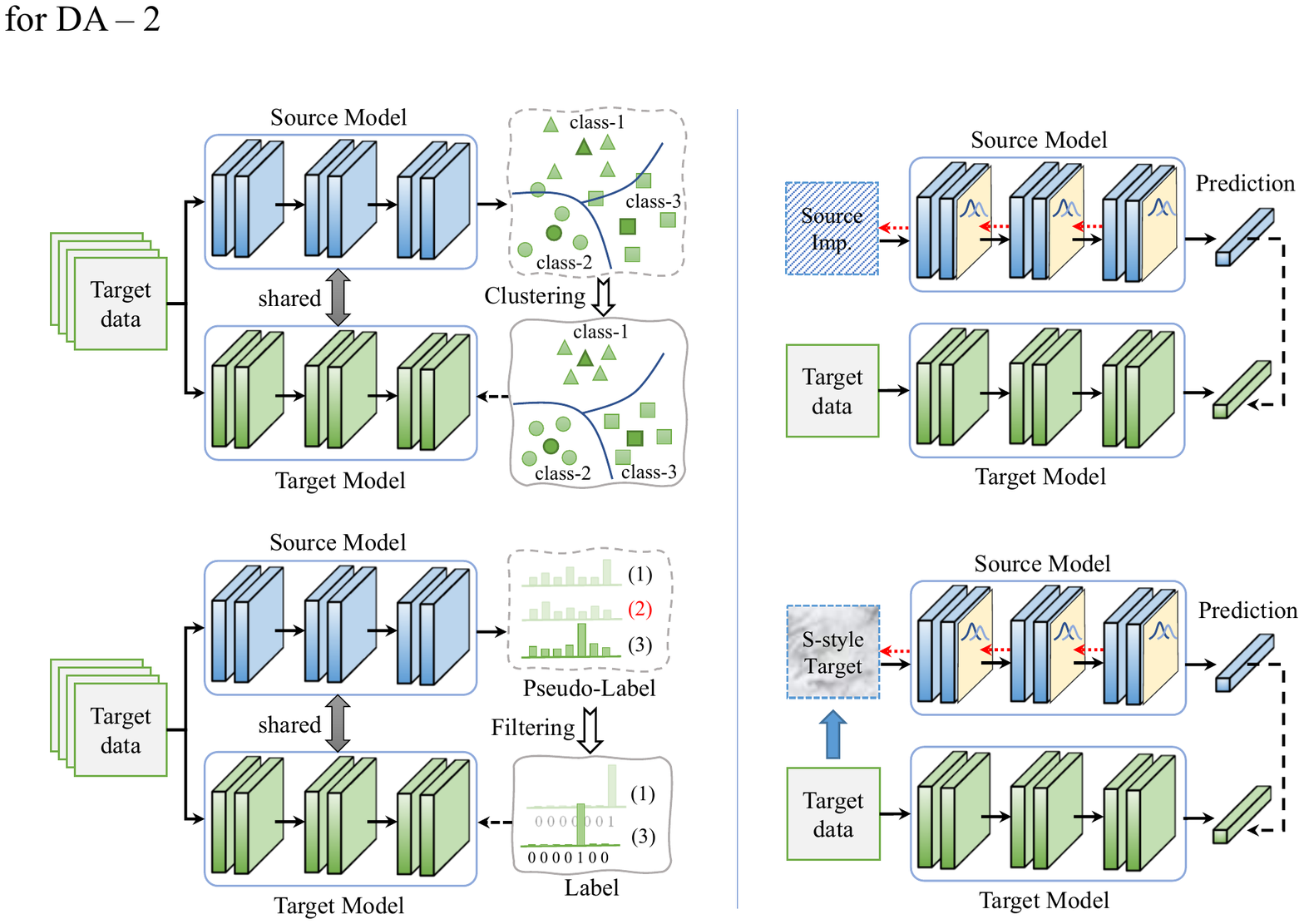}
}
\subfigure[Style Translation]{
  \includegraphics[width=0.7\linewidth]{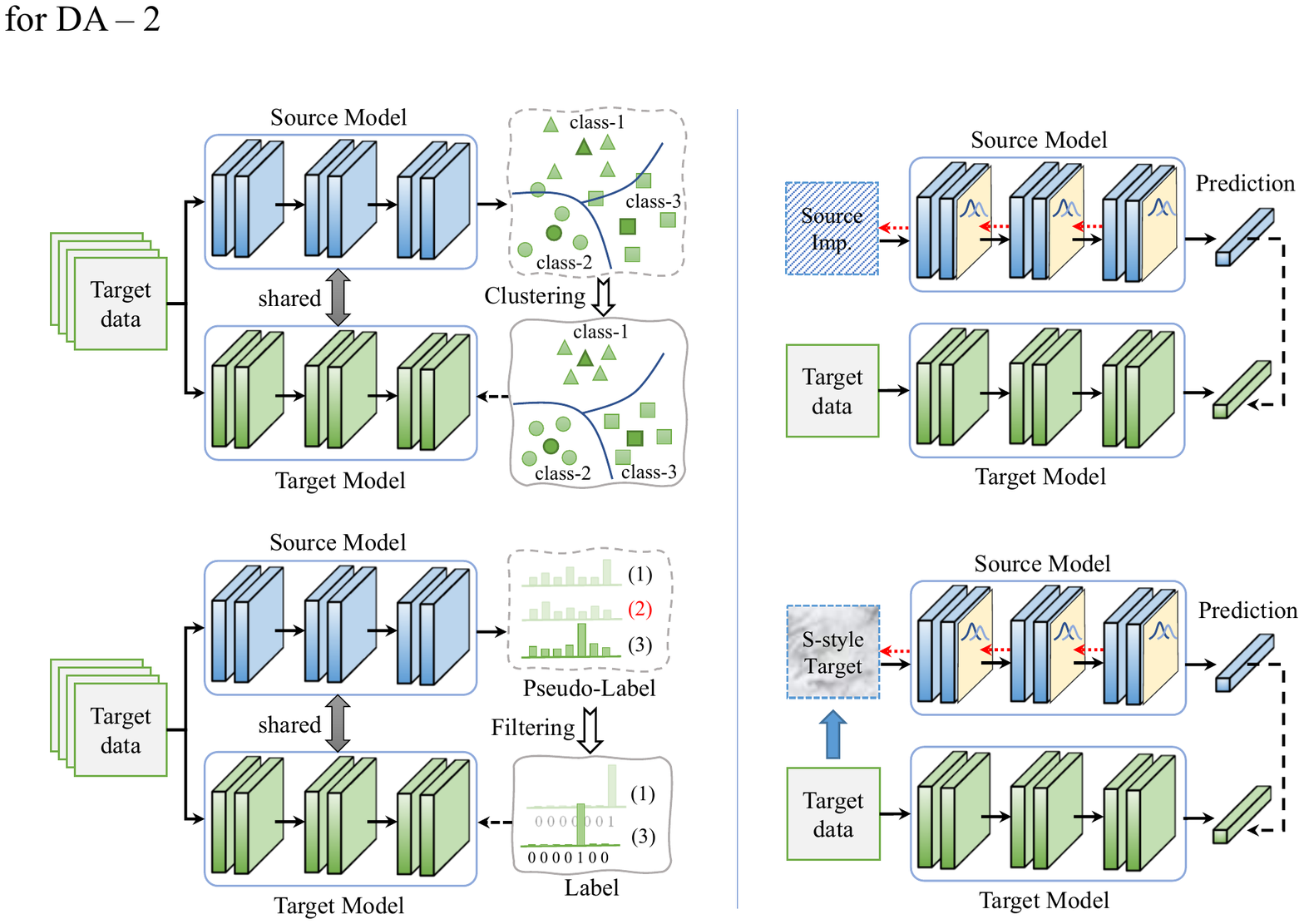}
}
\caption{SFDA with Virtual Source Knowledge Transfer: (a) source impression, (b) style translation. `Source Imp' represents the source impression, and `S-style target' means the source-style target image. }
\label{fig:sfda2}
\end{figure}

Inspired by DFKD, virtual source knowledge transfer methods attempt to synthesize some source impressions from the pre-trained source model, or translate the target data to the source style which can replace the source data.
They are illustrated in Fig.~\ref{fig:sfda2}. 
These synthetic source data can be utilized to distill source knowledge from the pre-trained model, which is then transferred to the target model for preventing source knowledge forgetting. We denote the synthetic source impressions or the source-style set as $\tilde{X}$. Then, the learning objective is formulated as
\begin{equation}
  \min_{\mathcal{P}, \mathcal{Q}} \mathcal{D}(\mathcal{P}(\tilde{X}), \mathcal{Q}(\tilde{X}) ) + \mathcal{L}_{\text{adap}}(\mathcal{Q}| X^t, \tilde{Y}^t) \,,
\end{equation}
where $\mathcal{D}$ measures the model discrepancy. $\tilde{Y}^t$ is the target pseudo-label set, calculated by Eqn.~\ref{eq:pseudo}. 

\subsubsection{Source Impression}

For tackling the issue of source data absence, we can synthesize the impressions of the source domain for joint training or knowledge transfer in adaptation. Therefore, most of the reviewed DFKD methods can be embedded in the DA framework to achieve this goal, as illustrated in Fig.~\ref{fig:sfda2}(a). 

The two studies~\cite{li2020model,kurmi2021domain} both introduce a generative adversarial framework to synthesize source impressions with the supervision of the source model prior and target images. In \cite{kurmi2021domain}, two discriminators are employed: the GAN discriminator is to differentiate real target data and synthetic source data, while the domain discriminator aims to promote the target model to obtain domain invariant features. And in \cite{li2020model}, only the former GAN discriminator is employed. Rather than generating source impressions, VDM-DA~\cite{tian2021vdm} models an intermediate virtual domain in the feature space based on a Gaussian mixture model (GMM).
It can work on 2D image-based and 3D point cloud-based cross-domain object recognition tasks. Similarly, CPGA~\cite{qiu2021source}, a domain generation and adaptation framework with contrastive prototypes, explores the boundary information of the source model and train a generator to synthesize source prototypes (not impressions).
They are used to align with the target data via contrastive learning. 
Towards domain generation, Frikha~\etal~\cite{frikha2021towards} attempt to extract the domain-specific knowledge from a pre-trained source model by generating domain-specific datasets. Following DeepDream~\cite{mordvintsev2015inceptionism} and DeepInversion~\cite{yin2020dreaming}, it generates an alternative dataset for each domain via noise optimization and then fuses the domain knowledge. 

\subsubsection{Style Translation} 

Unlike source impressions directly reconstructed from source models, the SFDA methods utilize style translation as shown in Fig.~\ref{fig:sfda2}(b).
Specifically, they transfer the target images to the source style to form pseudo source images, which can be used to distill the reliable target knowledge and predictions. 
In general, the style translation can be implemented by various data augmentations, transformations, such as brightness, contrast, \etc, or style transfer based on BN statistics.
Additionally, the style transfer could work on the image itself or the intermediate feature maps. 

Sahoo~\etal~\cite{sahoo2020unsupervised} propose a simple yet effective approach to adapt a source classifier to a target domain that varies from the source domain along natural axes, such as brightness and contrast. They leverage the softmax probabilities of the source classifier to learn transformations that bring target images closer to the source domain. In fact, this approach is limited to the intrinsic properties of images, and cannot support complex style shift between domains. Hou and Zheng~\cite{hou2020source} develop a generative approach to transfer the target images to source-style ones under the style and content constraints. Then they align the batch-wise feature statistics of generated images to that stored in batch normalization layers of a pre-trained model. 
From a perspective of measurement shift, Eastwood~\etal~\cite{eastwood2021source} address SFDA by adapting the feature extractor to realign the approximate feature distribution of the target data with the saved one on the source. 
However, in this way, the approximation of the feature distribution under the source data needs to be stored during pre-training, which could not be satisfied in some privacy-demanding scenarios. Instead of applying transformations or translation to the images, some researchers~\cite{ishii2021source,frikha2021towards} focus on adjusting the feature distribution at the BN layers, which implicitly achieves style transfer. Ishii and Sugiyama~\cite{ishii2021source} propose to achieve distributional alignment via matching the stored source BN statistics with those of target data input. They choose the KL divergence to measure the distribution discrepancy between domains when fine-tuning the trainable feature extractor, and adapt the classifier via information maximization. Inspired by the DFKD methods with adversarial exploration, Li~\etal~\cite{li2021divergence} propose an adversarial framework consisting of two alternative steps: generating adversarial examples and harnessing adversarial examples. In the first step, they employ a generator to synthesize smooth and diverse perturbations for training samples. In the second step, they learn from the generated examples to generalize better to the target domain. 
Recently, Yang~\etal~\cite{yang2022model} analyze the cross-domain representations in SFDA and derive a new theorem of generalization upper bound on the target prediction error, which is a general solution in this field. 

Compared with image synthesis from scratch/noise, the style translation methods are more promising for complex tasks and situations. 
This is because they can reuse some domain-invariant patterns in a target domain, which can reduce the complexity of pseudo source data generation.

\subsection{Application}
\label{sec:sfda_app}

We have thoroughly illustrated different kinds of SFDA methods towards classification, from principle to implementation. 
As a complement, this section briefly shows other applications of SFDA, \ie, semantic segmentation and object detection. 


\subsubsection{Semantic Segmentation}

\begin{table}[!t]
  \centering
  \caption{Source-free domain adaptation for semantic segmentation. `Filtering' is for pseudo-label filtering, `Impression' is for source impression, `Image ST / Aug' is for image style translation or augmentation, `Inter ST' is for intermediate style translation, and `GAN' is for checking whether generative adversarial framework is used. }
  \resizebox*{\linewidth}{!}{
    \begin{tabular}{c|cccc|c}
    \toprule
    Method & Filtering & Impression & Image ST / Aug & Inter ST & GAN \\
    \midrule
    Bateson~\etal~\cite{bateson2020source} & \cmark   &       &       &       & \xmark \\
    TENT~\cite{wang2021tent} &       &       &       & \cmark   & \xmark \\
    You~\etal~\cite{you2021domain} & \cmark   &       &       &       & \xmark \\
    Ye~\etal~\cite{ye2021source} & \cmark   &       &       &       & \cmark \\
    Teja \& Fleuret~\cite{fleuret2021uncertainty} &       &       & \cmark   &       & \xmark \\
    Liu~\etal~\cite{liu2021source} & \cmark   & \cmark   &       &       & \cmark \\
    SF-OCDA~\cite{zhao2021source} &       &       & \cmark   & \cmark   & \xmark \\
    Kundu~\etal~\cite{kundu2021generalize} & \cmark   &       & \cmark   &       & \xmark \\
    DAS$^3$~\cite{wang2021give} & \cmark   &    & \cmark   &       & \xmark \\
    PR-SFDA~\cite{luo2021exploiting} & \cmark   &       &       &       & \xmark \\
    S4T~\cite{prabhu2021s4t}  & \cmark   &       & \cmark   &       & \xmark \\
    You~\etal~\cite{you2021testtime} &       &       &       & \cmark   & \xmark \\
    Hong~\etal~\cite{hong2021source} &     &       & \cmark  &    & \cmark \\
    Paul~\etal~\cite{paul2022unsupervised} & \cmark   &       & \cmark   &       & \xmark \\
    \bottomrule
    \end{tabular}
  }
  \label{tab:sfda_app1}
\end{table}

Semantic segmentation is a pixel-level prediction task that aims to assign a semantic category label to each pixel of an image~\cite{long2015fully,chen2017rethinking}. The prediction output by a segmentation model is a label map. Thus, the image-level pseudo-label clustering strategy is not suitable for this task. 
However, the pseudo label maps can be split into positive and negative parts according to the prediction confidence. 
Only the positive part is utilized to update the model by self-training. 
Based on this, the virtual source knowledge transfer methods can be easily extended to segmentation model adaptation. 

Tab.~\ref{tab:sfda_app1} lists dozens of SFDA applications on semantic segmentation. We can see that most methods~\cite{bateson2020source,you2021domain,ye2021source,fleuret2021uncertainty,liu2021source,kundu2021generalize,wang2021give,luo2021exploiting,prabhu2021s4t,paul2022unsupervised} follow a self-training paradigm based on pseudo-label filtering and information maximization, which is simple yet effective for segmentation. \cite{bateson2020source} is the pioneering study of SFDA for segmentation, but it utilizes the pre-stored meta-data of the source domain, which cannot satisfy the privacy demand. TENT~\cite{wang2021tent} innovates to simply fine-tune the normalization and transformation parameters in BN layers to adapt the target model, needing less computation but still improving more. You~\etal~\cite{you2021testtime} propose to calibrate BNS by mixing up the source and target statistics for both alleviating the domain shift and preserving the discriminative structures in test time. SF-OCDA~\cite{zhao2021source} presents the cross-patch style swap and photometric transformation to simulate the real-world style variation, which could promote the model performance in semantic segmentation. Apart from updating the model via entropy minimization on selected pixels, You~\etal~\cite{you2021domain} introduce the negative learning to infer which category the negative pixels do not belong to and which is more feasible. DAS$^3$~\cite{wang2021give} employs the ClassMix~\cite{olsson2021classmix} technique to incorporate the hard and easy splits for regularizing the semi-supervised learning. S4T~\cite{prabhu2021s4t} builds a selective self-training framework to reduce the inconsistency between different views of an image. 
\cite{liu2021source} and \cite{hong2021source} both employ a generator to generate pseudo-source images based on BNS to transfer source knowledge to a target model. 
Besides,\cite{liu2021source} and \cite{ye2021source} introduce a discriminator following the target model to minimize the shift between the positive and negative split.

\subsubsection{Object Detection}
Object detection is a longstanding challenge which aims to predict the bounding boxes and the corresponding classes of the objects in an image. The domain adaptive object detection task is supposed to align both the image and object levels. And the prediction maps of a detection model are also dividable, similar to segmentation.

SFOD~\cite{li2021free} firstly attempts to address SFDA in object detection via modeling it into a problem of learning with noisy labels. Due to the uneven quality of pseudo labels, it introduces a metric method (named self-entropy descent) to search an approximate threshold for filtering reliable pseudo labels. Xiong~\etal\cite{xiong2021source} add perturbations to the target data and construct a super target domain which assists to explore domain invariant and specific spaces. With three consistency regularizations, the mean teacher model can align the target domain to the domain-invariant space. 

Recently, several works~\cite{saltori2020sf,hegde2021uncertaintyaware,hegde2021attentive} for 3D detection arise. 
SF-UDA$^\text{3D}$~\cite{saltori2020sf} is the first SFDA framework to adapt the PointRCNN~\cite{shi2019pointrcnn} 3D detector to target domains. It scales the input target data to various size and selects the best scale according to the consistency in different size. 
This method actually takes a multi-scale transformation of the point-cloud data to get the most reliable inference.
But it faces a terrible waste of time. To tackle this issue, Hegde~\etal~\cite{hegde2021uncertaintyaware} introduce a mean teacher with Monte-Carlo dropout uncertainty to supervise the student with iteratively generated pseudo labels. To address the limitations of traditional feature aggregation methods for prototype computation in the presence of noisy labels, Hegde and Patel~\cite{hegde2021attentive} leverage a transformer module to compute the attentive class prototype and identify correct regions-of-interest for self-training. 

In addition, GCMT~\cite{liu2021graph} handles the SFDA issue on person re-identification (ReID) with a graph consistency mean-teaching algorithm. Specifically, it constructs a graph for each teacher network to describe the similarity relationships among training images, and then fuses these graphs to provide supervised signals for optimizing the student network.

\subsection{Discussion}

Currently, most of the SFDA methods are designed specific to classification, and only tiny of works focus on more complex or dense prediction tasks, such as segmentation, detection, and ReID. Due to the nature of image-level labeling, the adaptive classification models are easy to be trained and fine-tuned via pseudo-label self-supervision, especially the simple yet effective pseudo-label clustering algorithms. 
The pseudo-label filtering algorithms can be extended to other complex tasks thanks to the separability of the prediction maps. 
To a certain extent, the filtering based self-supervision has been the most popular solution in these dense prediction tasks, with the assistant of entropy maximization.
This is because it takes the advantage of the part target data with limited information of the source domain. 

The virtual source knowledge transfer methods pay attention to reconstructing the source data space or converting target images into source-style ones.
This makes it convenient for transferring valuable information to the target domain. 
The source impression algorithms synthesize images following the source distribution, which increases both the difficulty and flexibility. 
On the other hand, the style translation methods make full use of the content in target images.
They can reduce the computation of synthesis but limit the variety of pseudo source. 
In particularly, some researchers~\cite{sahoo2020unsupervised,wang2021tent,zhao2021source} only adjust the style or characteristics of target images by varying from the source domain along natural axes, which is not flexible. 
We argue that the style translation is more efficient than the source impression when the target domain is very related to the source domain. 

In brief, the core of SFDA is to distill valuable source knowledge with an accessible target set. 
The main challenges in SFDA are the domain shift and the incompletely overlapping representation space. But the good news is that the unlabeled target set is available for usage and usually related to the source domain. 

\section{Future Research Directions}
\label{sec:frd}

In this section, we outline several future research directions that deserve more research efforts for data-free knowledge transfer from the methodology and task perspectives.

\subsection{New Methodology} 

\textbf{More efficient data reconstruction.} For the methods based on data reconstruction, the quality of the synthetic samples determines the effort of the knowledge transfer phase, and the efficiency of data generation algorithms affects the execution time of the pipeline. 
As such, exploring more efficient and reliable data reconstruction algorithms is one of the most promising research direction for data-free knowledge transfer. Some advanced  learning paradigms are supposed to be explored for valuable data generation, such as meta-learning~\cite{hospedales2020meta}, reinforcement learning~\cite{arulkumaran2017deep}, contrastive learning~\cite{jaiswal2021survey}, \etc. For SFDA in dense prediction tasks, how to take advantages of an unlabeled target set for source impression or style translation will draw more attention. Moreover, how to synthesize key samples according to the learning feedback is an interesting research direction. 

\textbf{Adaptive knowledge transfer.} Another key technique in data-free knowledge transfer is knowledge transfer strategies. The traditional data-driven KD methods~\cite{wang2021knowledge,gou2021knowledge} have been well studied. 
But in the data-free scenarios, the data-driven KD algorithms are not very suitable for noisy or fake data. Thus, apart from the data reconstruction, the adaptive knowledge transfer algorithms with different noisy data are urgently needed. The intermediate feature maps of the networks intend to express rich information. Given features from multiple layers, adaptive knowledge transfer approaches are more flexible and efficient, thanks to the consideration of different cases of layers.

\textbf{Ensemble/Federated learning.} Most of the exiting DFKD or SFDA frameworks only adopt one well-trained teacher/source model. However, ensemble learning~\cite{park2019feed,du2020agree} and multi-teacher distillation~\cite{liu2020adaptive} have shown great potential in traditional KD scenarios, while multi-source DA~\cite{peng2019moment,dong2021confident,ahmed2021unsupervised} has achieved significant success. On the one hand, multiple teachers can provide more consistent and reliable information about the original training data for DFKD. On the other hand, the target model can utilize different but comprehensive domain-related knowledge from multi-source models, which helps SFDA. In addition, federated learning is a promising research direction for privacy-demanding distributed applications, in which only the black-box models are accessible. 

\textbf{Transformer \& GNN.} Recently, some new network architectures like Transformer (especially vision Transformer)~\cite{han2020survey} and GNN~\cite{wu2020comprehensive}, have attracted researchers' interest and achieved remarkable performance. However, most of the existing DFKT methods are designed for convolutional neural networks (CNN). So how to develop specific algorithms for these new architectures remains to be addressed. Specifically, how to reconstruct data from pre-trained Transformers and GNNs, and how to transfer knowledge between them are the two major challenges. Different from 
CNN, Transformer works with image patches or tokens, while GNN works with data nodes and their edges.

\subsection{New Task Settings} 

\textbf{Task-agnostic DFKT.} As discussed above, most researchers focus on studying basic methods and applications of DFKT for simple classification tasks, but overlook other complex or task-agnostic learning scenarios, such as ReID, segmentation, super-resolution and so on. DFAD~\cite{fang2020data}, which only utilizes the predictions in an adversarial framework, and TENT~\cite{wang2021tent}, which simply adjusts the BN statistics parameters, point out the research directions of task-agnostic DFKD and SFDA, respectively. Besides, they also pose the challenge, \ie, simple and generic design for limited cases. 

\textbf{Robust DFKT.} Real-world samples generally suffer image noise or perturbation~\cite{yu2020label}, which require the robustness of networks in deployment. Most data-free methods, however, assume the original training data and the test data are clean, ignoring the model robustness on test data or a target domain. Since many data-free methods have tried to reconstruct the data and gained good performance, a straightforward solution is to combine robustness learning algorithms with alternative data in knowledge transfer. Inspired by this, robust target adaptation in the absence of source data, also deserves research efforts. 

\textbf{Sequential tasks.} Apart from single image or text data, the sequential data like videos or sentences, also faces the challenge of data-free knowledge transfer. The biggest problem is the generation of sequential data (especially videos), which is generally variable and with rich content. 

\textbf{Learning in the wild.} In the absence of original training data or source data, we can collect any available or related data on the web, including texts, images, videos, \etc.
They could be further filtered for the knowledge transfer. Chen~\etal~\cite{chen2021learning} have made a preliminary exploration of this research direction. 
Compared with data reconstruction based on pre-trained models, collecting related data in the wild is more convenient, and can avoid the difficulty of complex pattern synthesis.

\section{Conclusion}
\label{sec:con}

In this paper, we present a comprehensive and structured survey of data-free knowledge transfer. A hierarchical taxonomy and a formal definition of DFKT are provided at first. Then we discuss data-free knowledge distillation and source-free domain adaptation, and connect them under a unified DFKT framework. The two data-free aspects are categorized by their data reconstruction algorithms and knowledge transfer strategies, respectively. For each sub-category of approaches, we present detailed formal and visual description. To understand the impact of DFKD and SFDA, we further present their different applications. 
In the end of corresponding sections, a discussion of the advantages and challenges are provided.
Finally, we analyze the future research directions to benefit the community and attract more researchers' attention. Due to the growing concerns about data privacy and copyright, this hot topic has profound significance to promote the deployment and practical application of deep learning models. We hope our work can help readers have a better understanding of the research status and the research ideas.


%


\ifCLASSOPTIONcaptionsoff
  \newpage
\fi


\bibliographystyle{IEEEtran}
\bibliography{IEEEfull, ref} 

\begin{IEEEbiography}
  [{\includegraphics[width=1in,height=1.25in,clip]{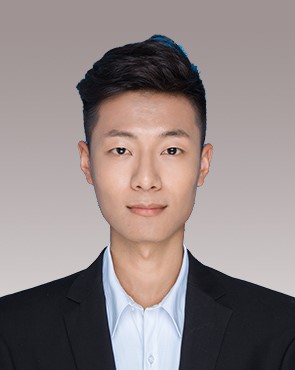}}]
  {Yuang Liu}
  is working toward the Ph.D. degree in the School of Computer Science and Technology, East China Normal University, Shanghai, China. He received his B.E. degree with the major of software engineering in Wuhan University of Science and Technology, in 2019. His research interests include computer vision, model compression and knowledge distillation. He has been invited as a reviewer for top international conferences and journals, including CVPR, ICCV, and Neurocomputing. 
\end{IEEEbiography}

\begin{IEEEbiography}
  [{\includegraphics[width=1in,height=1.25in,clip]{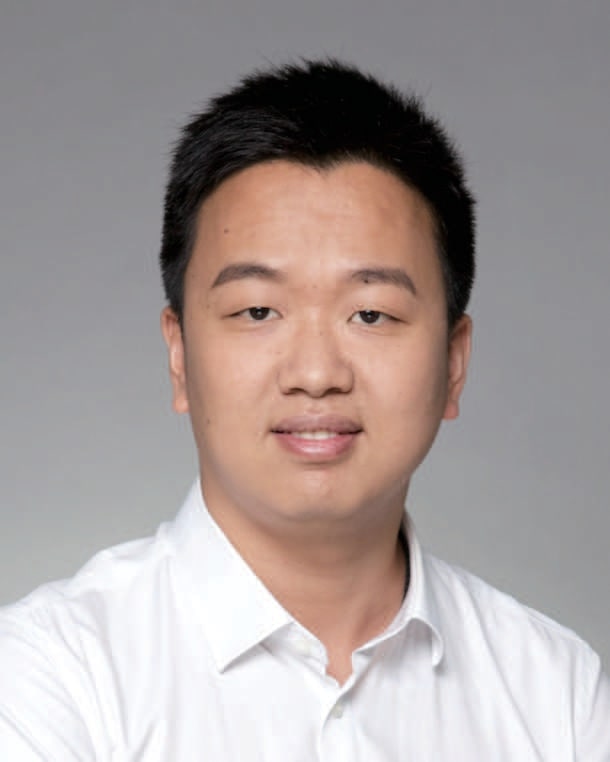}}]
  {Wei Zhang}
  received his Ph.D. degree in computer science and technology from Tsinghua University, Beijing, China, in 2016. He is currently an associate researcher in the School of Computer Science and Technology, East China Normal University, Shanghai, China. His research interests mainly include data mining and machine learning applications, especially focusing on user generated data.
\end{IEEEbiography}

\begin{IEEEbiography}
  [{\includegraphics[width=1in,height=1.25in,clip]{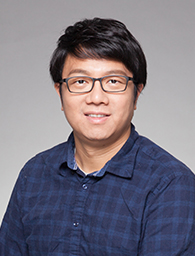}}]
  {Jun Wang}
  received the Ph.D. degree in electrical engineering from Columbia University, New York, NY, USA, in 2011. Currently, he is a professor at the School of Computer Science and Technology, East China Normal University and an adjunct faculty member of Columbia University. From 2010 to 2014, he was a research staff member at IBM T. J. Watson Research Center, Yorktown Heights, NY, USA. His research interests include machine learning, data mining, mobile intelligence, and computer vision. 
\end{IEEEbiography}

\begin{IEEEbiography}
  [{\includegraphics[width=1in,height=1.25in,clip]{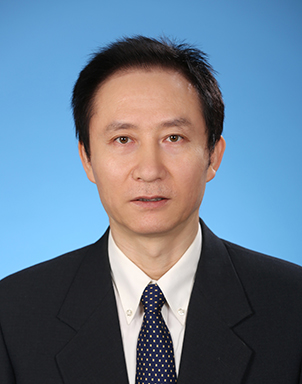}}]
  {Jianyong Wang (FIEEE)} received the Ph.D. degree in computer science from the Institute of Computing Technology, Chinese Academy of Sciences, in 1999. He is currently a professor in the Department of Computer Science and Technology, Tsinghua University, Beijing, China, and also with the Jiangsu Collaborative Innovation Center for Language Ability, Jiangsu Normal University, Xuzhou, China. His research interests mainly include data mining and Web information management. He has co-authored more than 60 papers in some leading international conferences and some top international journals. He is serving or has served as a PC member for some leading international conferences, such as SIGKDD, VLDB, ICDE, WWW, and an associate editor of the IEEE Transactions on Knowledge and Data Engineering and the ACM Transactions on Knowledge Discovery from Data. 
\end{IEEEbiography}




\end{document}